\pdfoutput=1

\documentclass{article}

    \PassOptionsToPackage{numbers, compress}{natbib}

    \usepackage[preprint]{neurips_2024}

\usepackage[utf8]{inputenc} 
\usepackage[T1]{fontenc}    
\usepackage{hyperref}       
\usepackage{url}            
\usepackage{booktabs}       
\usepackage{amsfonts}       
\usepackage{nicefrac}       
\usepackage{microtype}      
\usepackage{xcolor}         
\usepackage{amsmath}
\usepackage{multirow}
\usepackage{multicol}
\usepackage{algorithm}
\usepackage{algorithmic}
\usepackage{tcolorbox}
\usepackage{colortbl}
\usepackage{longtable}
\usepackage{mdframed}
\usepackage{inconsolata}
\usepackage{color}
\usepackage{xspace}

\usepackage{CJK}
\usepackage{svg}

\newcommand{\openbook}[1]{\texttt{\textcolor{green!50!black}{#1}}\xspace}
\newcommand{\closedbook}[1]{\texttt{\textcolor{purple!70!black}{#1}}\xspace}

\newcommand{\green}[1]{\textcolor{green!50!black}{#1}\xspace}
\newcommand{\purple}[1]{\textcolor{purple!70!black}{#1}\xspace}

\definecolor{darkgreen}{rgb}{0.0, 0.5, 0.0}
\definecolor{darkgray}{gray}{0.4}
\definecolor{maroon}{rgb}{0.5, 0.0, 0.0}
\definecolor{navy}{rgb}{0.0, 0.0, 0.5}
\definecolor{teal}{rgb}{0.0, 0.5, 0.5}

\title{LLMs Could Autonomously Learn Without External Supervision}

\author{Ke Ji$^{1,2\dagger}$, Junying Chen$^{1,2\dagger}$, Anningzhe Gao$^{1*}$, \textbf{Wenya Xie}$^{1,2}$ 
\\ \textbf{Xiang Wan}$^{1,2}$ \textbf{Benyou Wang}$^{1,2}$\thanks{Benyou and Anningzhe are the corresponding authors with email: \textit{wangbenyou}@cuhk.edu.cn.
The first two authors contributed to this work equally.} \\
$^1$ Shenzhen Research Institute of Big Data \\
$^2$ The Chinese University of Hong Kong, Shenzhen \\
\url{https://github.com/FreedomIntelligence/Autonomous_Learning}
\\
}

\begin{document}

\maketitle

\begin{abstract}

In the quest for super-human performance, Large Language Models (LLMs) have traditionally been tethered to human-annotated datasets and predefined training objectives—a process that is both labor-intensive and inherently limited. This paper presents a transformative approach: Autonomous Learning for LLMs, a self-sufficient learning paradigm that frees models from the constraints of human supervision. This method endows LLMs with the ability to self-educate through direct interaction with text, akin to a human reading and comprehending literature. Our approach eliminates the reliance on annotated data, fostering an Autonomous Learning environment where the model independently identifies and reinforces its knowledge gaps. Empirical results from our comprehensive experiments, which utilized a diverse array of learning materials and were evaluated against standard public quizzes, reveal that Autonomous Learning outstrips the performance of both Pre-training and Supervised Fine-Tuning (SFT), as well as retrieval-augmented methods. These findings underscore the potential of Autonomous Learning to not only enhance the efficiency and effectiveness of LLM training but also to pave the way for the development of more advanced, self-reliant AI systems.

\end{abstract}


\newcommand{\benyou}[1]{{ \color{blue} [Benyou says ``#1'']}}
\newcommand{\junying}[1]{{ \color{teal} [Junying says ``#1'']}}
\newcommand{\junyingzh}[1]{{ \color{teal} [Junying says ``\begin{CJK}{UTF8}{gbsn}\small #1\end{CJK}'']}}
\newcommand{\zhihong}[1]{{ \color{blue} [Zhihong says ``#1'']}}
\newcommand{\guiming}[1]{{ \color{blue} [Guiming says ``#1'']}}
\newcommand{\jike}[1]{{ \color{blue} [Jike says ``#1'']}}

\definecolor{outerboxcolor}{gray}{0.90} 
\definecolor{innerboxcolor}{rgb}{1,1,1}

\section{Introduction}

Large language models (LLMs)~\cite{kenton2019bert,qiu2020pre,han2021pre,touvron2023llama2} could learn from unsupervised corpora, supervised instruction data, and preference data (or reward models) according to a pre-defined and static learning objective.
Initially, pre-training allows these models to learn from vast amounts of human-generated text~\cite{ben2006analysis,pan2009survey,ganin2015unsupervised,long2016unsupervised}. 
This is followed by Supervised Fine-Tuning (SFT)~\cite{ben2006analysis,pan2009survey,ganin2015unsupervised,long2016unsupervised}, where models learn from human-annotated instruction data.
The third phase involves Reinforcement Learning from Human Feedback (RLHF)~\cite{christiano2017deep,stiennon2020learning,ouyang2022training,glaese2022improving,bai2022training}, where models are trained on human preference annotations.

These methods could be regarded as \textit{passive} learning strategies, where models passively absorb provided information without genuinely learning, not to mention consciously monitoring learning behaviors through self-reflection.
In real-world scenarios, humans demonstrate the capacity for Autonomous Learning, such as self-education through reading books or independent research of scientific papers.
Most human learning processes are subjective and require minimal guidance, exhibiting strong autonomous characteristics. 
This mode of learning is starkly different from the prescriptive nature of pre-training or SFT, which meticulously shapes the output of each token and symbol in the model. 
Additional, in the transition from weak to strong capabilities, the quality of human-annotated data has its limitations. 
In the future, the development of super-human LLMs cannot rely solely on human annotations but also requires Autonomous Learning by the LLMs themselves.

To mimic human learning, it reminds us to use Autonomous Learning, an ideal approach to human education. According to~\cite{little2002autonomy}, it is not merely a teaching method; hence, it does not involve teachers dictating behaviors for students to replicate. 
In ~\cite{holec1979autonomy}, the authors define Autonomous Learning as the capacity of learners to direct their own learning, implying their responsibility in shaping various aspects of the learning process. 
This includes critical thinking, planning, evaluating, and reflecting on learning, with learners actively monitoring the entire process~\cite{benson2013teaching}. 
Therefore, autonomous learners are reflective individuals who consciously strive to comprehend what, why, and how they learning~\cite{little1996politics}.
Consequently, while Autonomous Learning is considered an ideal approach, modern LLM training methods emphasize reliance on human-annotated data and predefined objectives, hindering learners' ability to autonomously monitor their learning process.

This inspirs us to adopt  \textbf{\textsc{Autonomous Learning}} for LLMs. 
The core idea is to enable LLMs to learn autonomously, without human involvement. 
In the context of Autonomous Learning, the only prerequisites are the Language Learning Model (LLM) itself and the learning resources, such as books or documents.
The process mimics how a person learns from a book: reading to understand and closing the book to recall and identify areas that require further study to reinforce knowledge. This approach boasts several unique advantages:

\begin{enumerate}
    \item  
    \textbf{Autonomous Learning.} 
    Unlike passive methods, Autonomous Learning involves the model actively engaging with and understanding the material, identifying areas for improvement, and reinforcing its knowledge—emulating the human process of self-improvement through learning.
    \item 
    \textbf{Dispensing with the need for annotations.} 
    As the model undertakes its own learning journey, human intervention becomes unnecessary. 
    The model is fed learning materials such as books, papers, or large corpora—and it dynamically improves itself without the need for annotated data from human, GPT-4 and others.
\end{enumerate}

To assess the efficacy of this learning method, we have set up experiments with learning materials of varying scales, such as books (10K paragraphs), domain-specific documents (100K paragraphs), and Wikipedia (1000K paragraphs), along with corresponding public quizzes (OpenBookQA, MedQA, TriviaQA, etc.) to evaluate the learning outcomes. Our experiments demonstrate that Autonomous Learning significantly outperforms pre-training and human-annotated SFT methods. Remarkably, Autonomous Learning also surpasses retrieval-augmented techniques (RAG), suggesting that a model that has diligently 'studied' could outperform one that has 'open-book' access but no review. Our findings confirm that Autonomous Learning is a more effective learning method, and its independence from annotations and human involvement significantly reduces the complexity and effort involved in model training.

The main contributions of this paper are listed as follows:
\begin{itemize}
\item We introduce \textbf{{Autonomous Learning}} for LLMs, a novel training paradigm that enables models to self-learn without human intervention or other stronger AI, mirroring the natural learning processes of humans.
\item We demonstrate that Autonomous Learning eliminates the need for human-annotated data, allowing models to actively engage with and understand learning materials, thereby fostering a dynamic and self-improving learning process.
\item Through rigorous experimentation using varied learning materials and corresponding public quizzes, we provide empirical evidence that Autonomous Learning significantly outperforms traditional pre-training and SFT methods, as well as retrieval-augmented techniques (RAG).
\end{itemize}

\section{Conceptualization: Autonomous Learning}

\subsection{Problem Statement}

We define a straightforward learning objective: Given a corpus \(\mathcal{D} = \{d_1, d_2, \ldots, d_n\}\) consisting of \(n\) documents, and a LLM \(\Phi_\theta\) with parameters \(\theta\), the goal is to enable \(\Phi_\theta\) to effectively learn from this corpus. The effectiveness of this learning can be evaluated using benchmarks related to \(\mathcal{D}\). This process is akin to a person studying a textbook for a course and then being assessed through course exams to gauge their understanding.

\subsection{Existing Learning for LLMs}
The training methods for LLMs can essentially be seen as human learning strategies.
The details of current methods closed to this paper is shown in Appendix~\ref{appendix:related_work}.

\paragraph{Pre-training}
Pre-training is a popular unsupervised training method, and its training objective can be formalized as follows:
\begin{equation}
\mathcal{L}_{\text{PT}}(\mathcal{D}) = -\sum_{d \in \mathcal{D}} \sum_{u \in d} \log P(u_i|u_{1},...,u_{i-1};\theta)
\end{equation}
where \(u_i\) represents the \(i\)-th token in \(d\). The pre-training objective is to maximize the prediction probability of each token. This can be likened to rote memorization in human learning, focusing on repetition rather than understanding.

\paragraph{Supervised Fine-Tuning (SFT)}
SFT is a common supervised learning method. Due to the high annotation costs of SFT, it is typically used for fine-tuning on downstream tasks or instruction fine-tuning. Its training objective is:
\begin{equation} 
\mathcal{L}_{\text{SFT}}(\mathcal{D},\mathrm{E}) = -\sum_{d \in \mathcal{D},\ (q,a) = \mathrm{E}(d)} \log P(a|q;\theta)
\end{equation}
where \(q\) and \(a\) represent the input and output respectively, and \(E\) denotes external annotation sources, such as humans or other LLMs. The goal of SFT is to learn to answer \(a\) given the question \(q\). SFT relies on external sources \(E\) to provide the external understanding of \(d\). This is analogous to learning from a teacher's guidance, where the teacher's understanding (\(E\)) is imparted to the student \(\Phi_\theta\).

\subsection{Autonomous Learning}
Pre-training and SFT are limited because they involve models passively absorbing information without truly understanding or self-monitoring their learning processes. In contrast, humans excel at Autonomous Learning, such as self-education through reading or independent research, which requires minimal guidance and involves actively understanding and reinforcing new knowledge. As human-annotated data has its limitations, future development of super-human LLMs will require models to adopt similar autonomous learning strategies, going beyond the prescriptive and passive nature of pre-training and SFT.

\paragraph{Autonomous Learning (AL)}  To this end, we propose Autonomous Learning for LLMs, enabling them to learn autonomously like humans. Unlike previous methods, all learning content in Autonomous Learning is self-generated (self-understanding).  Autonomous Learning simulates this human learning process in two stages. The first stage is \textbf{open-book} learning, with the learning objective:
\begin{equation}
    \mathcal{L}_{\text{OpenBook}}(\mathcal{D}) = - \sum_{d \in \mathcal{D},\ (q,a_o) = \Phi_\theta(\text{Prompt}(d))} \log P(a_o|q;\theta)
\end{equation}
Here, \(q\) and \(a_o\) represent the model \(\Phi_\theta\)'s self-generated understanding of \(d\), where \text{Prompt} is the prompt that helps the model understand document \(d\). During this process, \(\Phi_\theta\) thoroughly absorbs the book's content. The second stage of Autonomous Learning is \textbf{closed-book} gap-filling, with the learning objective:
\begin{equation}
    \mathcal{L}_{\text{ClosedBook}}(\mathcal{D}) = 
    -\sum_{ d \in \mathcal{D},\ (q,a_o) = \Phi_\theta(\text{Prompt}(d))} \log
    \sigma (P(a_o|q;\theta)-P(\Phi_\theta(q)|q;\theta))
\end{equation}
Here, \(\Phi_\theta\) represents the model's reasoning. In this process, the model aligns its closed-book generated answers \(\Phi_\theta(q)\) with the open-book answers \(a_o\). Since the model has already acquired a preliminary understanding in the first stage, the second stage aims to identify and reinforce knowledge gaps and areas where learning is insufficient.

\section{Methodology: Autonomous Learning}

\begin{figure}[!ht]
\centering
\includegraphics[width=0.9\textwidth]{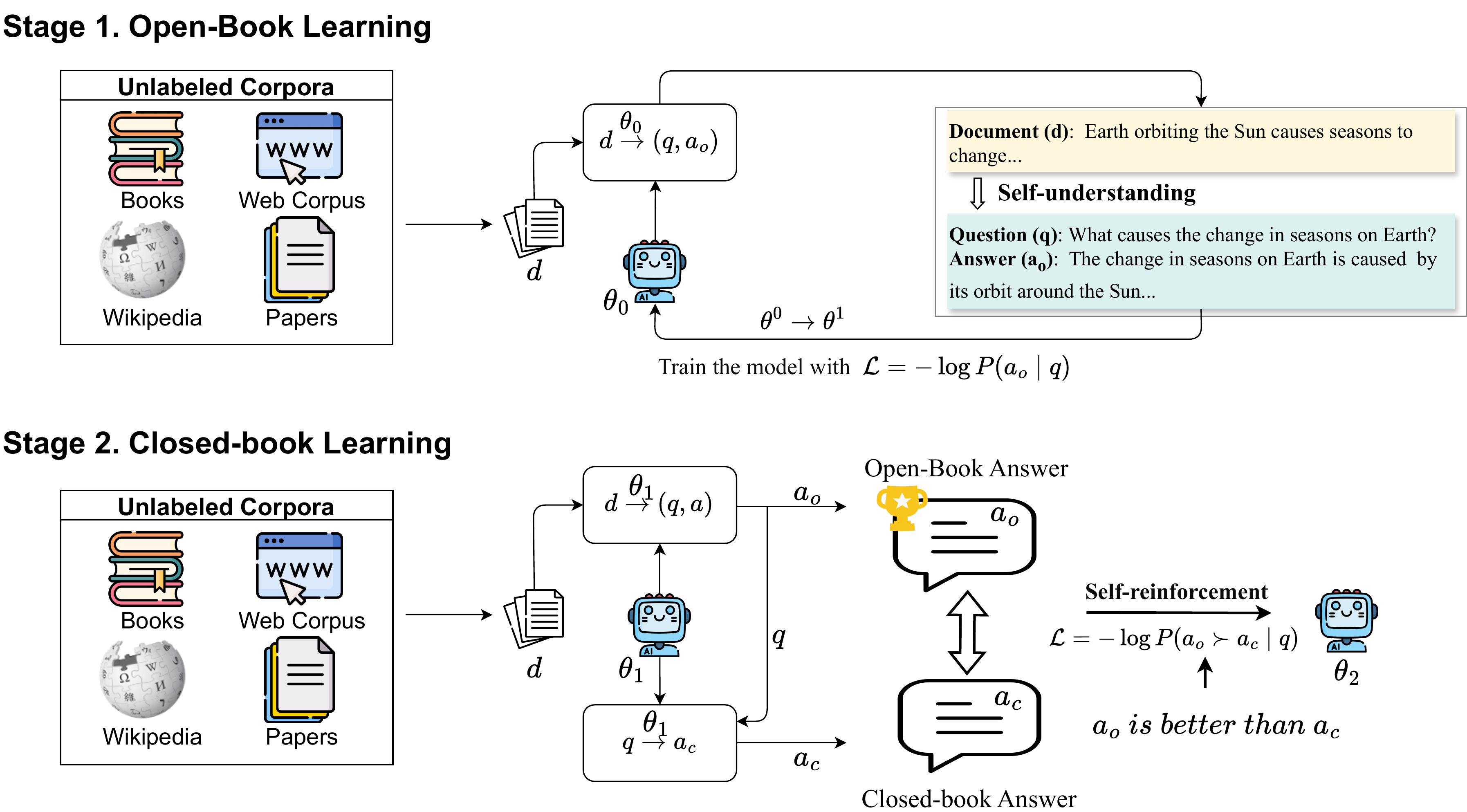}
\caption{
An ideal learning system should learn autonomously to determine \textit{what to learn}, \textit{how to learn} and \textit{why to learn}.
}
\label{fig:AL_Overview}
\end{figure}

\label{sec:method}
In this section, we provide a detailed implementation of our proposed Autonomous Learning. The overview of our Autonomous Learning framework is shown in Figure~\ref{fig:AL_Overview}. Autonomous Learning enables LLMs to learn in a manner similar to humans. This process consists of two stages:  \textbf{Stage 1. Open-book learning:} The model comprehends and absorbs the textual information. \textbf{Stage 2. Closed-book learning:}  The model recalls the content from the first stage, reinforcing and consolidating the learned material. The entire algorithm flow of Autonomous Learning is shown in Algorithm \ref{alg:1}.

\begin{algorithm}[ht!]
\small
\caption{The algorithm of Autonomous Learning}
\label{alg:overall_framework_tmp}
{\bf Input:}
$\Phi_{\theta^0}$, $\mathcal{D}$ \\
{\bf Output:}
$\Phi_{\theta^2}$
\begin{algorithmic}[1]
\STATE { \color{teal} \quad // Stage 1. Open-Book Learning} 
\STATE $\theta^1 \gets \theta^0$
\FOR{document $d$ in $\mathcal{D}$}
    \STATE $(q, a_o) \gets \Phi_{\theta^{0}}(\text{Prompt}(d))  $ {\color{teal} \quad // Comprehending document} 
    \STATE $\ell_1 \gets - \log P(a_o | q;\theta^1)$  
    \STATE $\theta^1 \gets \text{UpdateParameters}(\ell_1,\theta^1)$ {\color{teal} \quad // Absorbing document}  
\ENDFOR \\
\STATE {\color{teal} \quad // Stage 2. Close-Book Learning} 
\STATE $\theta^2 \gets \theta^1$
\FOR{document $d$ in $\mathcal{D}$}
    \STATE $(q, a_o) \gets \Phi_{\theta^{1}}(\text{Prompt}(d))$
    \STATE $a_c \gets \Phi_{\theta^{1}}(q)$
    \STATE $\ell_2 \gets - \log \sigma \left(\beta \log \frac{\pi_{\theta^2}(a_o\mid q)}{\pi_{\theta^1}(a_o\mid q)} - \beta \log \frac{\pi_{\theta^2}({a_c}\mid q)}{\pi_{\theta^1}({a_c}\mid q)}\right)$ {\color{teal} \quad // Self-reinforcement} 
    \STATE $\theta^2 \gets \text{UpdateParameters}(\ell_2,\theta^2)$ 
\ENDFOR
\RETURN $\Phi_{\theta^{2}}$
\end{algorithmic}
\label{alg:1}
\end{algorithm}

\subsection{Stage 1. Open-Book Learning}
Open-book learning simulates the process of studying a book, where we comprehend and absorb its content. The initialization model for Autonomous Learning is a LLM with comprehension abilities, denoted as $\Phi_{\theta^0}$. Given a document $d$ to be learned, $\Phi_{\theta^0}$ first comprehends $d$ before learning it. This comprehension process can be seen as reading the document and converting it into questions and answers (QA), which can be formalized as:
\begin{equation}
(q, a_o) = \Phi_{\theta^{0}}(\text{Prompt}(d))    
\end{equation}

Here, $q$ and $a_o$ represent the questions and answers generated from the document $d$,, and $\text{Prompt}$ refers to the prompt used, as illustrated in Figure \ref{alprompt}. For LLMs that cannot follow the prompts, we provide few-shot examples to enable $\Phi_\theta^0$ to have comprehension abilities, as shown in Appendix B. 
In AL, $\Phi_\theta^0$ first learns from all documents $d \in D$. For documents that are too long, we split them into multiple paragraphs for learning, as detailed in Appendix A. The objective of open-book learning is:
\begin{equation}
\mathcal{L}_{\text{OpenBook}}(d) = -\log P(a_o|q;\theta^1)
\end{equation}
Thus, we obtain the model $\Phi_\theta^1$ after the first stage of learning.

\newtcolorbox{alprompt}[1]{
        boxrule = 1pt,
        fontupper = \small\tt,
        fonttitle = \bf\color{black},
        arc = 2pt,
        rounded corners,
        colframe = black,
        colbacktitle = white!97!yellow,
        colback = white!97!yellow,
        title = #1,
}

\begin{figure}[h]
\centering\small
\label{fig:al_prompt}
\begin{alprompt}{\centering \textit{The prompt for document comprehension}}
Please create a question that closely aligns with the provided article. Ensure that the <question> does not explicitly reference the text. You may incorporate specific scenarios or contexts in the <question>, allowing the <text> to serve as a comprehensive and precise answer, at the same time, you need to generate an <answer> for the generated <question>. You can refer to the content of the article to answer, but your answer cannot reveal that you have referred to this article. Please output according to the template: '<question>: ... <answer>: ...'

<document>: \colorbox{outerboxcolor}{\color{teal} [domain-specific document]}\\
<question>: \\
<answer>: 
\end{alprompt}
\caption{\label{alprompt} The prompt for document comprehension. \colorbox{outerboxcolor}{\color{teal}[domain-specific document]} indicates the document $d$ to be learned.}
\end{figure}

\subsection{Stage 2. Closed-book Learning}

The model $\Phi_{\theta^1}$ from the first stage can be thought of as a person who has read a book once. Human learning generally involves a review process to consolidate knowledge, similar to studying for an exam at the end of a course. In this process, we usually close the book and recall previously learned content to enhance memory. For the LLM, the second stage involves having the model $\Phi_{\theta^1}$ recall the learned content without referring to the document, thereby reinforcing the knowledge. We first have the model generate QA pairs based on $d$:
\begin{equation}
(q, a_o) = \Phi_{\theta^{1}}(\text{Prompt}(d))
\end{equation}
Note that the questions $q$ generated for the same $d$ vary. For the abstracted questions $q$ from $d$, Autonomous Learning has the model answer them with the book closed:
\begin{equation}
a_c = \Phi_{\theta^{1}}(q)
\end{equation}
where $a_c$ represents the closed-book answers. This gives us a pair $(a_o, a_c)$. We aim to have the model's closed-book answers $\Phi_{\theta^{1}}(q)$ approximate $a_o$ as closely as possible. To achieve this, we use a Direct Preference Optimization (DPO) strategy to help the LLM improve the review process. The advantage of DPO is its ability to quickly approximate the correct answers in the presence of biased data. The DPO learning strategy is as follows:
\begin{equation}
\label{eq:cloesd_book_obj}
\mathcal{L}_\text{CloseBook}(d) = - \log \sigma \left(\beta \log \frac{\pi_{\theta^2}(a_o\mid q)}{\pi_{\theta^1}(a_o\mid q)} - \beta \log \frac{\pi_{\theta^2}({a_c}\mid q)}{\pi_{\theta^1}({a_c}\mid q)}\right)
\end{equation}

where $\pi_{\theta^1}({a_c}\mid q)$ represents the probability of model $\Phi_{\theta^{1}}$ generating $a_c$ given $q$. In this process, Autonomous Learning treats the open-book answer $a_o$ as the positive answer and the closed-book answer $a_c$ as the negative answer, achieving a self-reinforcing process.

\subsection{The Benefits of Autonomous Learning}
Autonomous Learning enables LLMs to understand and learn on their own,  like humans. This offers several advantages:
 
    \textbf{I: Self-Learning in a Loop~} 
    Unlike passive learning (i.e., pre-training or supervised instruction learning), Autonomous learning enables the model to engage in self-learning in a loop. 
    This means the model repeatedly generates content, evaluates its own outputs, adjusts and improves based on the evaluation results, and then generates new content. This loop allows the model to continuously self-optimize and enhance its performance, similar to how humans learn through constant reflection and improvement.

    \textbf{II: No Need for External Annotations~} 
       Autonomous Learning does not require external annotations, unlike SFT that depends on them. The model's self-understanding drives its learning journey, rendering human intervention unnecessary. As LLMs advance to super-human capabilities, the model's self-derived understanding could surpass that of human-generated content. Consequently, Autonomous Learning becomes increasingly effective as the model improves.

    \textbf{III: Simplified Learning Process~} 
   Autonomous Learning eliminates the need for data processing, data cleaning, or managing data mixing ratios. The only requirement is to provide the corpus \( D \) to the LLM. 
   This greatly simplifies the training process of LLMs and ensures high-quality learning outcomes.

\section{Experiments}

We evaluate our Autonomous Learning (AL) framework across various domains, including commonsense reasoning and domain-specific QA. We compare AL to traditional knowledge injection methods, assess its scalability with different dataset sizes, and its efficacy in specialized fields like medicine. We also analyze the impact of Open-Book and Closed-Book learning on performance, and test AL's consistency across different initial models. 
\subsection{Target Domain With Various Scales and Downstream Tasks}

To highlight the superiority of our method, we consider the size of the knowledge corpus included in each dataset when selecting them, which varies from 1K to 1M.
We train on knowledge corpus and test on multiple downstream tasks corresponding to these specific corpus.
The details of our used benchmark is shown in Appendix~\ref{appendix:dataset}.

In all instances, we adopt a prompted zero-shot setup, wherein models are directed to address each task using natural language instructions without any accompanying contextual examples.
We choose the more challenging zero-shot setup as we are interested in seeing whether Autonomous Learning works in precisely those cases where a AI system does not specify in advance which instruction should be used in which way for solving a specific problem.
In fact, we let the model directly complete downstream tasks to test the model's ability to master knowledge in a specific domain.
We use standard greedy decoding.
The statistics of these datasets can be found in Table~\ref{tab:DatasetStatistics}.
All tasks are measured by accuracy.
For tasks under Wiki, we use the reference answers after minor normalization operations mentioned in ~\cite{chen2017reading,lee2019latent}.
\begin{table}[!ht]
\caption{The statistical information of the used benchmark.}
\footnotesize
\centering
\setlength{\tabcolsep}{1.0mm}
\begin{tabular}{lccccccccc}
\toprule

\multirow{2}{*}{\textbf{Dataset}}  
	
&\multicolumn{1}{c}{\textbf{Commonsense}} &\multicolumn{3}{c}{\textbf{Medical}}  &\multicolumn{5}{c}{\textbf{Wiki}} \\

\cmidrule(lr){2-2} \cmidrule(lr){3-5} \cmidrule(lr){6-10}
 &OpenBookQA &CNPLE &MedQA-en &MedQA-cn  &NQ &TriviaQA & WebQA &TREC &SQuAD \\ 
 \midrule
Train &\multicolumn{1}{c|}{4957} & - &  10178& 27400  & \multicolumn{1}{|c}{78168} &78785 &3417 & 1353 & 78713 \\
Dev &\multicolumn{1}{c|}{500} & - &1272  &3425   & \multicolumn{1}{|c}{8757} &8837 &361 & 133 & 8886 \\
Test  &\multicolumn{1}{c|}{500} & 960 &1273 &3426 & \multicolumn{1}{|c}{3610} &11313 &2032 &694 &10570   \\
\midrule
\rowcolor{gray!17}\multicolumn{10}{c}{ \textbf{Number of documents for each dataset, ranging from 1K to 1M}}    \\ 
Documents &\multicolumn{1}{c|}{1326} &87096 &156960 &163843 &\multicolumn{5}{|c}{1M} \\ 
\bottomrule
\end{tabular}
\label{tab:DatasetStatistics}
\end{table}

\subsection{Experiments Setup}

\paragraph{Experimental settings.}

Our research concentrates on unsupervised adaptation scenarios, utilizing Autonomous Learning on an unlabeled target domain corpus to train and enhance an initial model. We hypothesize that a robust model will demonstrate effective generalization and high performance on the target domain's test sets. Our ultimate aim is to transform this model into a domain-specific expert and an instruction model for chat applications, thereby demonstrating the potential of Autonomous Learning in model enhancement and domain-specific adaptation.

\paragraph{Base Model.}
We use the meta-llama/Llama-2-7b-chat-hf for experiments, which we call it as \textbf{initial model} in our experiments. 
This model originate from HuggingFace~\footnote{\url{https://huggingface.co/}}.

\subsection{Baselines.}
\label{sec:baselines}
To compare with other baselines broadly, we replicate the setups used by prior work and reuse their reported numbers whenever possible.
We note that for most tasks, our goal is not to compete with the state-of-the-art (SOTA) because: 1) for tasks like Multi-Choice and open domain question answering, SOTA models are trained specifically for the corresponding training sets; and 2) SOTA methods often use additional corpora for pretraining that may lead to data contamination, which could confound our domain adaptation studies.
We consider the following baselines for our experiments.

\textbf{Pre-training}: Following the traditional pre-training paradigms proposed in \cite{kenton2019bert,radford2018improving,tay2022ul2}, we implement a vanilla pre-training method, which lets the model be pretrained by performing conventional autoregressive language modeling on a given corpus.

\textbf{Supervised Fine-tuning (SFT)}: We implement a SFT~\cite{ouyang2022training} method named InstructGPT to perform unsupervised domain adaptation.
InstructGPT utilizes a substantial amount of manually annotated data, which incurs significant costs. To avoid hallucinations, we use a stronger model to build instructions for 10\% of the documents, while for the remaining documents, we use the model itself to build instructions.

\textbf{Retrieval Augmented Generation (RAG)}: 
RAG first performs a retrieval step to identify the most relevant document fragments. 
These retrieved document fragments are then fed into the generative model to serve as the context for generating responses. 
The generative model constructs an answer based on this additional context and the patterns it has learned internally.
For each question, we retrieve four documents.

\textbf{Imbalanced Learning (IL)}: 
We implement active bias~\cite{chang2017active}, a widely used method in imbalanced learning that directly adjust the weights of examples based on the variance in their predictive distributions.
We perform IL on pre-training and supervised fine-tuning, and we get 'pre-training + IL' and 'supervised fine-tuning + IL'.

\begin{table*}[h]
\caption{Results on Common sense and Medical corpora.
The best performances are highlighted in \textbf{bold}, while sub-optimal ones are marked with \underline{underline}. 
}
\small
\centering
\begin{tabular}{lcccccccccc} 
\toprule
 \multirow{2}{*}{\textbf{Model}}  
&\multicolumn{1}{c}{\textbf{Commonsense}} 
&\multicolumn{3}{c}{\textbf{Medcical}} 
& \multirow{2}{*}{\textbf{Avg Acc.}}  
\\	
\cmidrule(lr){2-2} \cmidrule(lr){3-5}
&OpenBookQA  &MedQA-cn &MedQA-en &CNPLE  \\ 
    \midrule[0.6pt]
    initial model &35.0 &26.2 &30.5 &19.3 &27.8 \\ \midrule
    \multicolumn{6}{l}{ \textbf{Passive methods}}     \\ 
    Pre-training &37.0 &42.6 &31.4 &30.4  &35.4\\
    Pre-training+IL &38.4 &41.8 &30.5 &27.6  &34.6\\
    RAG &38.4  &28.4 &26.2 &26.0 &29.8\\
    Supervised Fine-Tuning  &\underline{42.0} &52.4 &33.2  &41.8 &42.4 \\
    Supervised Fine-Tuning+IL &41.4 &\underline{53.3} &\underline{33.6} &\underline{42.4} &\underline{42.7}\\
    \midrule
    \multicolumn{6}{l}{ \textbf{Autonomous methods}}     \\ 
    \rowcolor{orange!17} Autonomous Learning (Ours) &\textbf{53.0} &\textbf{58.2} &\textbf{37.5} &\textbf{46.4}  &\textbf{48.8}\\
    \bottomrule
\end{tabular}
\label{tab:main_results_1}
\vspace{-3mm}
\end{table*}

\subsection{Scaling Laws Across Multi-Magnitude Corpora}
\label{sec:main_results}
As training in deep learning and large language models becomes increasingly expensive, neural scaling laws can ensure performance. Before training large language models with hundreds of millions of parameters on massive corpora, we initially train models on smaller-scale corpora and fit scaling laws for training on larger corpora.

Unlike previous work~\cite{henighan2020scaling,yang2023baichuan}, which typically fix the size of the corpus and vary the scale of model parameters to observe the effects on error, this paper's scaling laws focus more on the corpus. 
The aim is to demonstrate through experiments on scaling laws of corpora size that our method is universally effective across various scales of corpora.
As shown in Table~\ref{tab:main_results_1} and Table~\ref{tab:main_results_2},
the benchmark results demonstrate that the Autonomous Learning outperforms all the currently most popular knowledge learning paradigms across various document scales.
In specific domains such as Medical, the method described in this paper still shows significant improvements.

\begin{table*}[h]
\caption{Results on a large corpora with 1 million Wiki documents.}
	\centering

	\begin{tabular}{lccccccc} 
	\toprule
 \multirow{2}{*}{\textbf{Model}}  
	&\multicolumn{5}{c}{\textbf{Wiki}} &\multirow{2}{*}{\textbf{Avg Acc.}} 
 
 \\
        \cmidrule(lr){2-6}
        &\multicolumn{1}{c}{\textbf{NQ}}
	&\multicolumn{1}{c}{\textbf{TriviaQA}}
        &\multicolumn{1}{c}{\textbf{WQ}}
        &\multicolumn{1}{c}{\textbf{TREC}}
	&\multicolumn{1}{c}{\textbf{SQuAD}} 
        
 \\
 
        \midrule[0.6pt]
        initial model &32.3 &57.6 &50.1 &29.6 &22.3 &38.4 \\
        \hline
        \multicolumn{7}{l}{ \textbf{Passive methods}}     \\ 
        Pre-training  &33.5 &64.3 &50.3 &30.4 &22.6  &40.2\\
        Pre-training+IL  &34.1 &63.6 &50.3 &31.6 &21.3 &40.2 \\
        RAG &\textbf{43.6} &\textbf{71.5} &50.7 &28.7  &21.5  &\underline{43.2}\\
        Supervised Fine-Tuning  &37.5 &66.2 &\underline{52.2} &32.5 &\underline{23.4} &42.4\\
        Supervised Fine-Tuning+IL  &36.3 &68.5 &52.1  &\underline{33.1} &22.7 &42.5 \\
        \midrule
        \multicolumn{7}{l}{ \textbf{Autonomous methods}}     \\ 
        \rowcolor{orange!17} Autonomous Learning (Ours) &\underline{39.2} &\underline{69.1} &\textbf{54.4} &\textbf{35.9} &\textbf{24.5}  &\textbf{44.6}\\
        \bottomrule
	\end{tabular}
\label{tab:main_results_2}
\end{table*}

\begin{table}[b]
\caption{Ablation study.}
\centering
\begin{tabular}{@{}c|lccccc}
\toprule
&Ablation model& OpenBookQA & MedQA-cn &MedQA-en & CNPLE               \\ 
\midrule
 - & initial model     & 35.0   &26.2    &30.5   &19.3       \\
\midrule
i & open-book only    & {40.0} & {51.4} & {32.4} &{40.5}    \\
ii & closed-book only  & {44.4} &{52.6}   &{33.7} &{42.3}    \\
iii & closed-book $\rightarrow$ open-book & 48.4 & 54.3 & 35.2 &44.1  \\
iv & AL w/o $\purple{a_c}$ in closed-book & 33.6 & 25.4 & 28.3 &19.6  \\
v & AL w/o reference model & 51.2 & 56.1 & 35.6 &43.6  \\

\midrule
\rowcolor{orange!17} vi & open-book $\rightarrow$ closed-book (AL)  & \textbf{53.0} & \textbf{58.2} & \textbf{37.5} &\textbf{46.4}  \\

\bottomrule
\end{tabular}
\label{tab:ablation_study}
\vspace{-2mm}
\end{table}

\subsection{Ablation Study}
To better explore the impact of each part of our model, we conducted ablation studies and the results are shown in Table~\ref{tab:ablation_study}.

By analyzing the comprehensive ablation experiment settings, we can draw the following conclusions: 1) All ablation models except iv can improve the capabilities of the initial model. 2) Closed-book learning is better than open-book only models i.\\

Furthermore, we find that \textbf{ablation model iv yield results as expected, even lower than the initial model.} 
One possible explanation is that when removing all terms related to the closed-book answer $a_c$ from the learning objective Formula~\ref{eq:cloesd_book_obj} during the closed-book learning phase, the learning objective of closed-book learning approximates open-book learning. 
Consequently, training for more epochs leads to overfitting, thereby reducing effectiveness.
This finding highlights the effectiveness of AL, wherein self-reflective knowledge contrast further strengthens the model's ability to generalize knowledge.
The more detailed experimental results regarding the generalization performance of the Autonomous Learning in two stages are presented in Appendix~\ref{appendix:not_enough}. The experimental results indicate that, without the need for additional external annotations, Closed-Book learning can further enhance the knowledge generalization performance of existing fine-tuning paradigms.

Interestingly, when we directly perform closed-book learning (the ablation model iii), the performance has certain advantages compared to open-book learning, but this effect is still far lower than the complete Autonomous Learning model.
The reason may be due to the lack of learning of all documents by the model in the open-book learning stage. 
As a result, when closed-book learning is performed directly, although the model's learning method based on self-knowledge comparison can learn a certain amount of knowledge, it is still under-fitting.

\subsection{Effects on Various Models}
To highlight the scalability of our method, we deployed our experiments using Baichuan 2-Chat-7b as our initial model. 
The experimental results are shown in Table~\ref{tab:deploy_on_other_models}.

Baichuan 2-Chat-7b, after Autonomous Learning training, shows significant improvement, comparable to Llama 2-Chat-7b. 
Notably, its performance on Chinese datasets like MedQA-cn and CNPLE slightly surpasses that of Llama-2-7b-chat-hf. 
This could be due to Baichuan 2-Chat-7b's higher learning potential with Chinese data, as Llama-2-7b-chat-hf's training corpus is primarily in English.

\begin{table}[!t]
\centering
\caption
{Experiment of deploying Autonomous Learning (AL) on various LLMs as our initial models.}
\begin{tabular}{@{}lccccc}
\toprule
& OpenBookQA & MedQA-cn &MedQA-en & CNPLE   \\ 
\midrule
 initial model (Llama-2-7b-chat-hf) & 35.0 & 26.2 & 30.5 & 19.3  \\
\rowcolor{orange!17}\quad + AL &{53.0} &{58.2} &{37.5} &{46.4} \\  
\midrule
 initial model (Baichuan 2-Chat-7b) & 34.0 & 27.4 & 31.0 & 19.8  \\
\rowcolor{orange!17}\quad + AL & {52.4} & {59.4} & {37.1} &{47.1}   \\
\bottomrule
\end{tabular}

\label{tab:deploy_on_other_models}
\vspace{-3mm}
\end{table}

\subsection{Competitive Performance Achieved by Fewer Documents}
The Closed-Book phase of our approach aims to enhance the model's generalization of learned knowledge and can be seamlessly integrated into any model that has undergone the Open-Book learning phase to further enhance its learning effectiveness. To investigate the knowledge enhancement effects of our approach in the Closed-Book learning phase, we conducted an in-depth exploration of the relationship between model performance and the quantity of documents used for reinforced knowledge learning in this phase.

Table~\ref{tab:fewer_documents} illustrates the experimental results of our approach in the Closed-Book phase under different scales of document subsets. It can be observed that our approach in the Closed-Book phase demonstrates performance comparable to the full dataset when based on only 30\% of the documents. 
Additionally, when only 5\% of the documents are available, our approach rapidly enhances the model's generalization of knowledge, achieving performance on par with SFT.

This highlights the efficient utilization of documents by our approach, which can extract rich knowledge through self-learning even with a small number of documents, thereby enhancing the model's generalization of knowledge.

\begin{table}[h]
\caption
{Low-resource settings where it adopts  fewer documents in Autonomous Learning (AL).}
\centering
\begin{tabular}{@{}lccccc}
\toprule
& OpenBookQA & MedQA-cn &MedQA-en & CNPLE   \\ 
\midrule
initial model & 35.0 & 26.2 & 30.5 & 19.3  \\
SFT &42.0 & 50.3 & 33.0 & 40.8    \\
\midrule

AL with full documents & \textbf{53.0} & \textbf{58.2} & \textbf{37.5} &\textbf{46.4}   \\
\rowcolor{gray!17}\multicolumn{5}{l}{\textbf{AL with fewer document}} \\ 
\quad \# 30\% & {50.2} & {56.9} & {36.6} &{45.6}   \\
\quad \# 15\% & {44.2} & {52.4} & {35.3} &{43.3}   \\
\quad \# 5\% & {38.6} & {51.6} & {34.2} &{39.5}    \\

\bottomrule
\end{tabular}
\label{tab:fewer_documents}
\vspace{-3mm}
\end{table}

\section{Conclusion}
In this paper, we introduce and validate \textsc{Autonomous Learning} as a groundbreaking training paradigm for Large Language Models (LLMs). By enabling LLMs to self-educate through direct interaction with diverse textual materials, this approach not only mimics human learning processes but also significantly enhances the capabilities of LLMs beyond the constraints of traditional training methods reliant on human-annotated data. Our results show that this approach outperforms traditional methods like supervised pre-training, SFT, and RAG techniques, offering a more efficient path to advanced AI systems. This shift towards autonomous self-improvement in LLMs heralds a new era of sophisticated, self-reliant AI capable of continuous learning without human intervention.

\section*{Acknowledgement}
This work was supported by  the Shenzhen Science and Technology Program (JCYJ20220818103001002), Shenzhen Doctoral Startup Funding (RCBS20221008093330065), Tianyuan Fund for Mathematics of National Natural Science Foundation of China (NSFC) (12326608), Shenzhen Key Laboratory of Cross-Modal Cognitive Computing (grant number ZDSYS20230626091302006), and Shenzhen Stability Science Program 2023, Shenzhen Key Lab of Multi-Modal Cognitive Computing.

\clearpage

\clearpage

{
\bibliographystyle{unsrt}
\bibliography{custom}
}

\clearpage
\appendix

\section{Related Work}
\label{appendix:related_work}
\subsection{Unsupervised Domain Adaptation (UDA)} 
Traditional UDA methodologies encompass Pseudo-labeling \cite{ye2020feature}, the Pivot-based approach \cite{pan2010cross}, and adversarial neural networks \cite{ganin2016domain}. 
Due to success of self-supervised learning paradigm's ability to utilize large-scale unlabeled data, 
pre-trained language models~\cite{kenton2019bert,qiu2020pre,han2021pre,radford2019language} based on self-supervision have become the standard paradigm in unsupervised domain adaptation.
Recently, Adaptive pre-training on domain-specific datasets has emerged as a potent adaptation strategy, exemplified by BioBERT \cite{lee2020biobert}, a specialized variant of BERT. 
AdaptaBERT \cite{han2019unsupervised} introduces a secondary phase of unsupervised pre-training for BERT to facilitate unsupervised domain adaptation. 
UDALM \cite{karouzos2021udalm} advocates for a mixed multi-task loss framework for simultaneous classification and masked language modeling. 
Embarrassingly \cite{chronopoulou2019embarrassingly} employs an auxiliary language modeling loss to mitigate catastrophic forgetting during transfer learning.
Although simplified, this line of methods exhibit \textbf{poor generalization} to unseen tasks and demonstrate subpar instruction-following capabilities, making it difficult to achieve satisfactory interactions in variable task scenarios.

\subsection{Supervised Fine-Tuning (SFT)}
It has been demonstrated that SFT language models on a collection of datasets expressed in instruction form~\cite{longpre2023flan,touvron2023llama2,yang2023baichuan} can improve model performance and generalization to unseen tasks, resulting many instruction-based supervised fine-tuning methods~\cite{chung2024scaling,touvron2023llama1,touvron2023llama2} have been introduced.
Additionally, a serious of work are proposed to focus on how to adapt large language model to a specific domain, such as Chatlaw~\cite{cui2023chatlaw}, Investlm~\cite{yang2023investlm}, Chatharuhi~\cite{li2023chatharuhi} and HuotuoGPT-II~\cite{chen2023huatuogpt}.

Although exciting, the SFT method of instructing the model \textit{what to learn} relies heavily on a large amount of high-quality annotations from humans, GPT-4~\cite{openai2023gpt4}, or other sources,
posing a formidable barrier to the scalability of instruction tuning practices for larger corpora in the future.

\subsection{Imbalanced Learning (IL)}
Existing imbalanced learning research can be broadly divided into five orthogonal categories:
(i)~\textbf{Loss-oriented methods} employs strategies such as reweighting~\cite{huang2016learning,chang2017active,cao2019learning,cui2019class,wang2024unified} to make the model pay more attention to minority classes during optimization;
(ii)~\textbf{Data-oriented methods} involve resampling to balance the training set, such as over-sampling minority classes~\cite{kubat1997addressing,chawla2002smote} and under-sampling majority classes~\cite{mani2003knn,drummond2003c4};
Despite their intuitiveness, over-sampling can cause overfitting, and under-sampling might reduce the information available for model training, as noted in~\cite{johnson2019survey}.
(iii)~\textbf{Post-hoc methods} adjust the model's outputs during the test phase after standard training. These include calibrating decisions using data priors~\cite{collell2016reviving}, balancing classifier weights with $\tau$-normalization~\cite{kang2019decoupling}, and adjusting logits for balanced accuracy, as practiced in~\cite{menon2020long};
(iv)~\textbf{Decoupling methods} apply a two-stage learning process, starting with conventional feature learning followed by retraining classifiers under a balanced label distribution ~\cite{kang2019decoupling} or implementing techniques like L2-normalization to stabilize classifier weights~\cite{alshammari2022long};
(v)~\textbf{Ensemble methods} merge insights from multiple models trained under different conditions. 
In ~\cite{cai2021ace}, the authors allocate overlapping yet distinct class splits among experts, promoting the acquisition of complementary knowledge by each expert.
Other strategies include employing dynamic routing to lessen model variance and bias~\cite{wang2020long}, effectively addressing performance issues in majority classes and shifting focus progressively to minority classes~\cite{zhou2020bbn}.

Despite their achievement, these methods have two main drawbacks:
\begin{itemize}
    \item Although intuitive, this line of approaches~\cite{huang2016learning,chang2017active,cao2019learning,cui2019class,wang2024unified} may send \textbf{incorrect signals} to the optimization process of \textit{how to learn}.
    \item Most of these methods involve \textbf{complex protocols}, making it difficult to deploy them in the knowledge learning process of large language models.
\end{itemize}

\section{Target Domain With Various Scales and Downstream Tasks}
\label{appendix:dataset}
Below we describe each domains and its corresponding downstream tasks.

\noindent\textbf{Commonsense:}
We choose a small-scale corpus dataset in the domain of common sense, OpenBookQA, which contains a corpus of 1,326 common sense entries to serve as reference knowledge for test data.
\begin{itemize}
    \item \textbf{OpenBookQA}~\cite{mihaylov2018can} comprises 5,957 multiple-choice questions, each offering four possible answers. 
    The dataset is combined with external fundamental scientific facts.
    To successfully answer these questions, one must have a comprehensive understanding of these fundamental scientific facts.
and its applications.
\end{itemize}

\noindent\textbf{Medical:}
We pick three widely used datasets in Medical domain.
Each dataset is accompanied by a medical textbook, which contains the knowledge required to answer the questions in the dataset. 
We split the textbook corpus into multiple documents, each containing no more than 512 tokens. After dividing the textbooks, the CNPLE, MedQA-en, and MedQA-cn datasets contain 87,096, 156,960, and 163,843 documents, respectively.
Please note that MedQA-cn and CNPLE are written in Chinese.
\begin{itemize}
    \item \textbf{MedQA-en}~\cite{jin2021disease} gathers questions from the National Medical Board Examinations of the USA. 
    MedQA presents a demanding benchmark because it incorporates diverse medical knowledge—including patient profiles, disease symptoms, and drug dosage requirements. 
    This variety requires contextual understanding for accurately answering the questions posed.
    \item \textbf{MedQA-cn}~\cite{jin2021disease} is also collected from the National Medical Board Examinations of the Mainland China.
    For both MedQA-en and MedQA-cn, we test them on the 4-option questions.
    \item \textbf{The 2023 Chinese National Pharmacist Licensure Examination (CNPLE)}~\cite{chen2023huatuogpt} is a fresh medical exams.
    Addressing data contamination in the training of Large Language Models (LLMs) is challenging, particularly when dealing with complex and vast datasets~\cite{ceval}. 
    To mitigate this issue, we use the 2023 Chinese National Pharmacist Licensure Examination, conducted on October 21, 2023, as our benchmark. The release date of this dataset is later than all the base and chat models we used, therefore it can prevent data leakage and ensure reliable evaluations.
    
\end{itemize}

\noindent\textbf{Wiki:}
We use the same five QA datasets and training/dev/testing splitting method as in previous work~\cite{lee2019latent}.
For datasets under this part, we train on the documents in Wiki corpus as their common corpus.
Here, we select a subset of the Wikipedia corpus that contains 1 million documents.
\begin{itemize}
    \item \textbf{Natural Questions (NQ)}~\cite{kwiatkowski2019natural} was designed for end-to-end question answering. The questions were mined from real Google search queries and the answers were spans in Wikipedia articles identified by annotators.
    \item \textbf{TriviaQA}~\cite{joshi-etal-2017-triviaqa} contains a set of trivia questions with answers that were originally scraped from the Web.
    \item \textbf{WebQuestions (WQ)}~\cite{berant2013semantic} consists of questions selected using Google Suggest API, where the answers are entities in Freebase.
    \item \textbf{CuratedTREC (TREC)}~\cite{baudivs2015modeling} sources questions from TREC QA tracks as well as various Web sources and is intended for open-domain QA from unstructured corpora.
    \item \textbf{SQuAD v1.1}~\cite{rajpurkar2016squad} is a popular benchmark dataset for reading comprehension.
    Annotators were presented with a Wikipedia paragraph, and asked to write questions that could be answered from the given text.
\end{itemize}

\section{Hyperparameters of Autonomous Learning}
The training hyperparameters of Autonomous Learning on different datasets are reported in Table~\ref{tab:hyper_al}. 
For all of the hyperparameters, we directly use the same value across all datasets.
The training was conducted on a GPU server with 8 NVIDIA A100 GPU cards.

\begin{table}[h]
\caption{The hyperparameters used for Our Autonomous Learning on all benchmark.}
\centering
\small
\begin{tabular}{l|l|ccccc}
\hline
\toprule
&Hyperparameters & OpenBookQA & CNPLE & MedQA-en & MedQA-en & wiki  \\
\midrule
                        & Optimizer  & \multicolumn{5}{c}{AdamW}    \\
                        & Warmup Ratio  & \multicolumn{5}{c}{0.1}   \\
                        & Learning Rate  & \multicolumn{5}{c}{2e-5}   \\
Open-Book Stage         & LR Schedule & \multicolumn{5}{c}{cosine}  \\
                        & Batch Size & \multicolumn{5}{c}{8}        \\
                        & Max Length & \multicolumn{5}{c}{2048}     \\
                        & \# Epoch  & \multicolumn{5}{|c}{3}    \\
\midrule

                        & Optimizer         & \multicolumn{5}{c}{Rmsprop} \\
                        & Warmup Ratio      & \multicolumn{5}{c}{0.2} \\
                        & Learning Rate  & \multicolumn{5}{c}{5e-7}   \\
                        & LR Schedule       & \multicolumn{5}{c}{Linear} \\
Closed-Book Stage         & Batch Size        & \multicolumn{5}{c}{8} \\
                        & Max Length        & \multicolumn{5}{c}{2048}  \\
                        
                        & DPO beta          & \multicolumn{5}{c}{0.01} \\
                        & \# Epoch          & \multicolumn{5}{c}{3} \\
\bottomrule
\end{tabular}
\end{table}
\label{tab:hyper_al}

\section{Naive Empirical Risk Minimization is Not Enough}
\label{appendix:not_enough}   
\begin{figure*}[ht]
    \begin{minipage}[t]{0.33\linewidth}
        \centering
        \includegraphics[width=\textwidth]{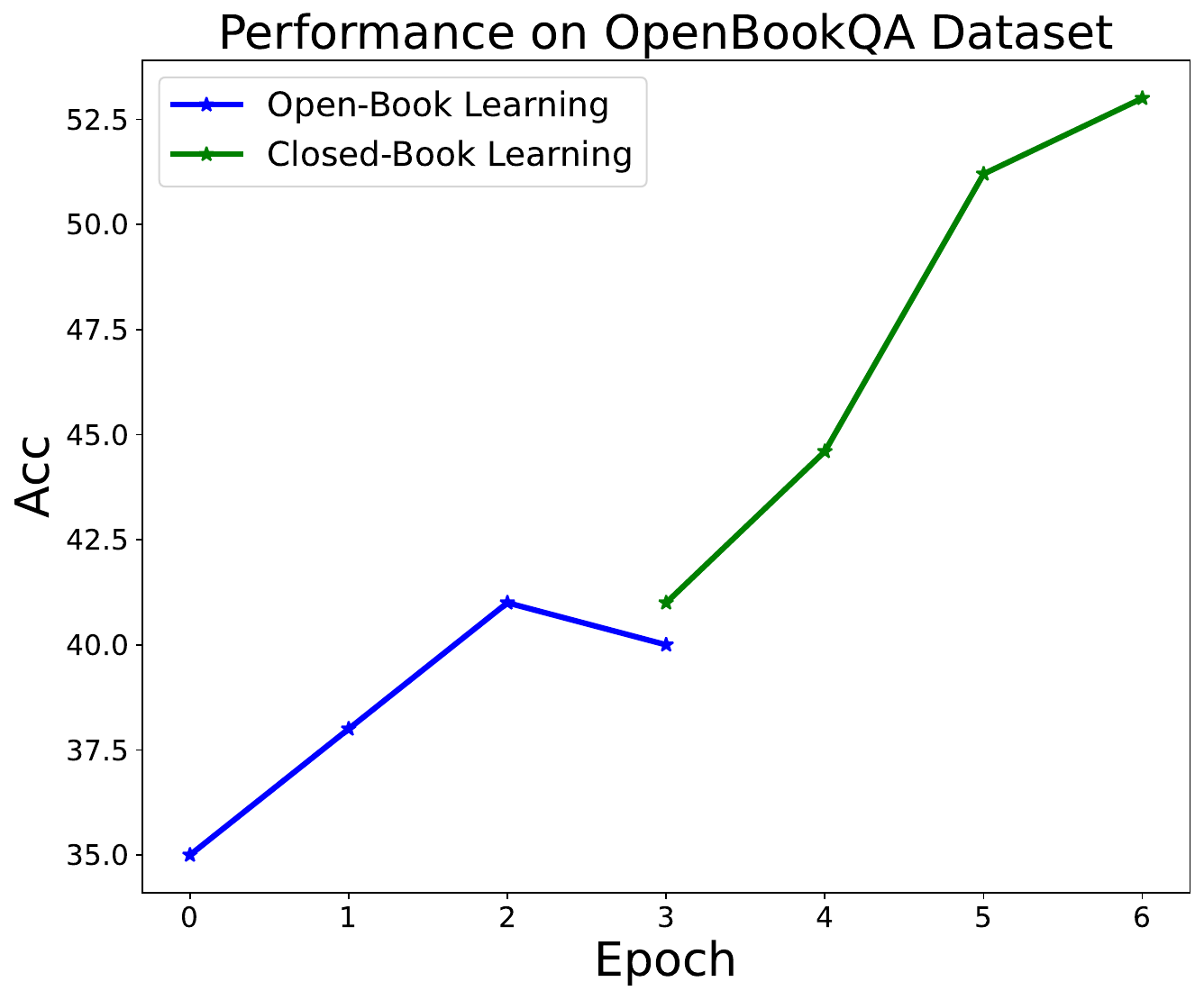}
        \centerline{(a)}
    \end{minipage}%
    \begin{minipage}[t]{0.33\linewidth}
        \centering
        \includegraphics[width=\textwidth]{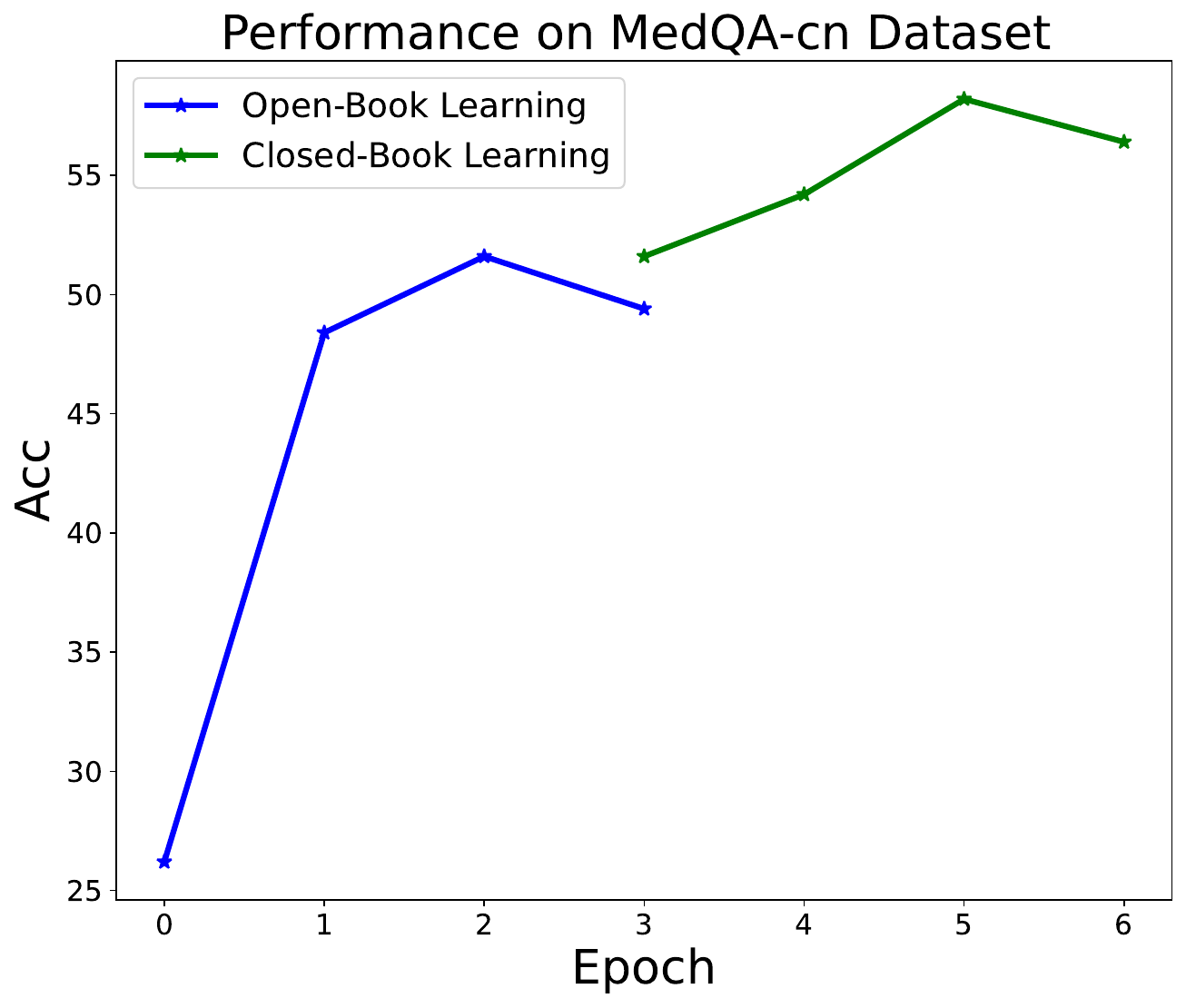}
        \centerline{(b)}
    \end{minipage}
    \begin{minipage}[t]{0.33\linewidth}
        \centering
        \includegraphics[width=\textwidth]{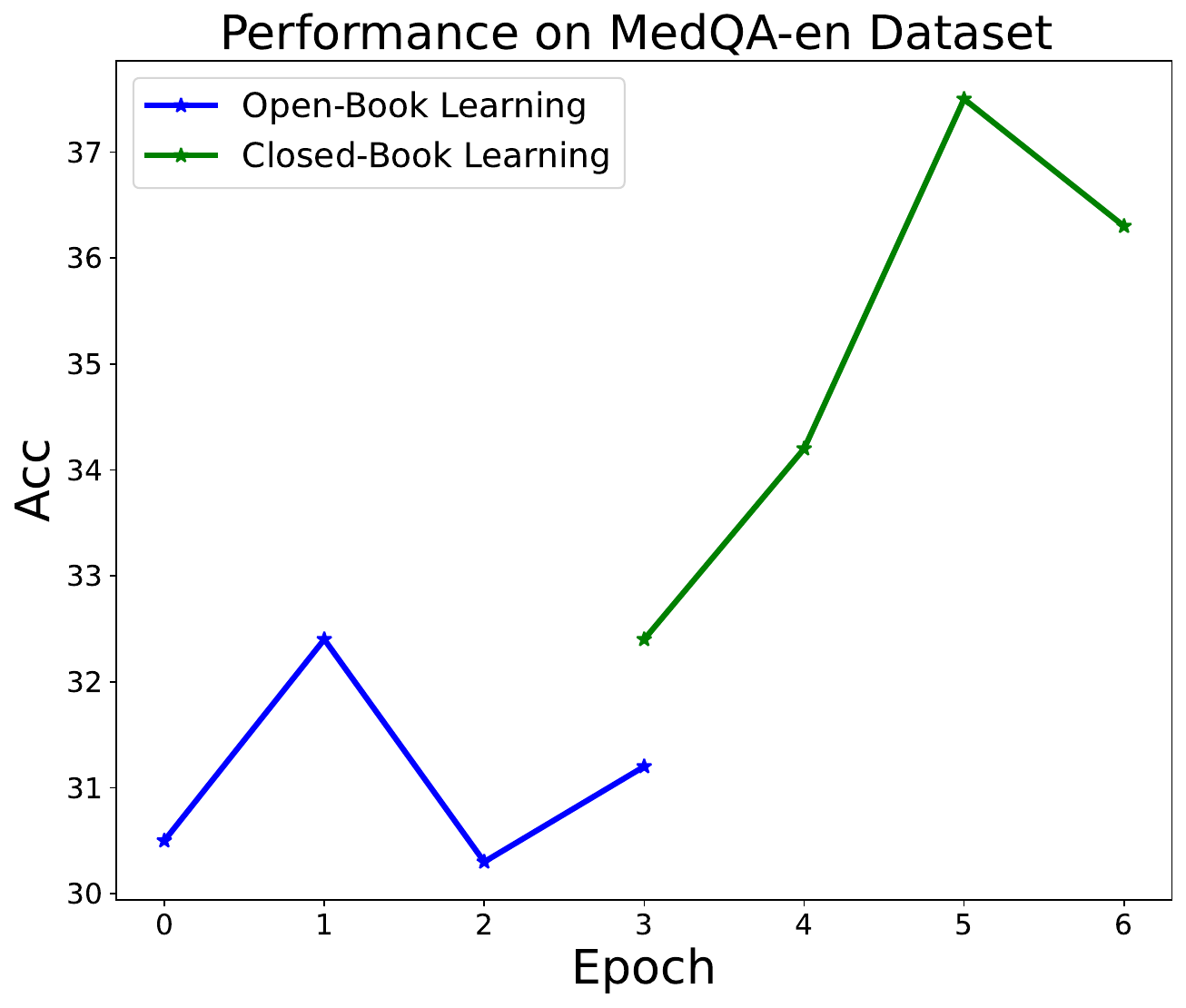}
        \centerline{(c)}
    \end{minipage}
    \begin{minipage}[t]{0.33\linewidth}
        \centering
        \includegraphics[width=\textwidth]{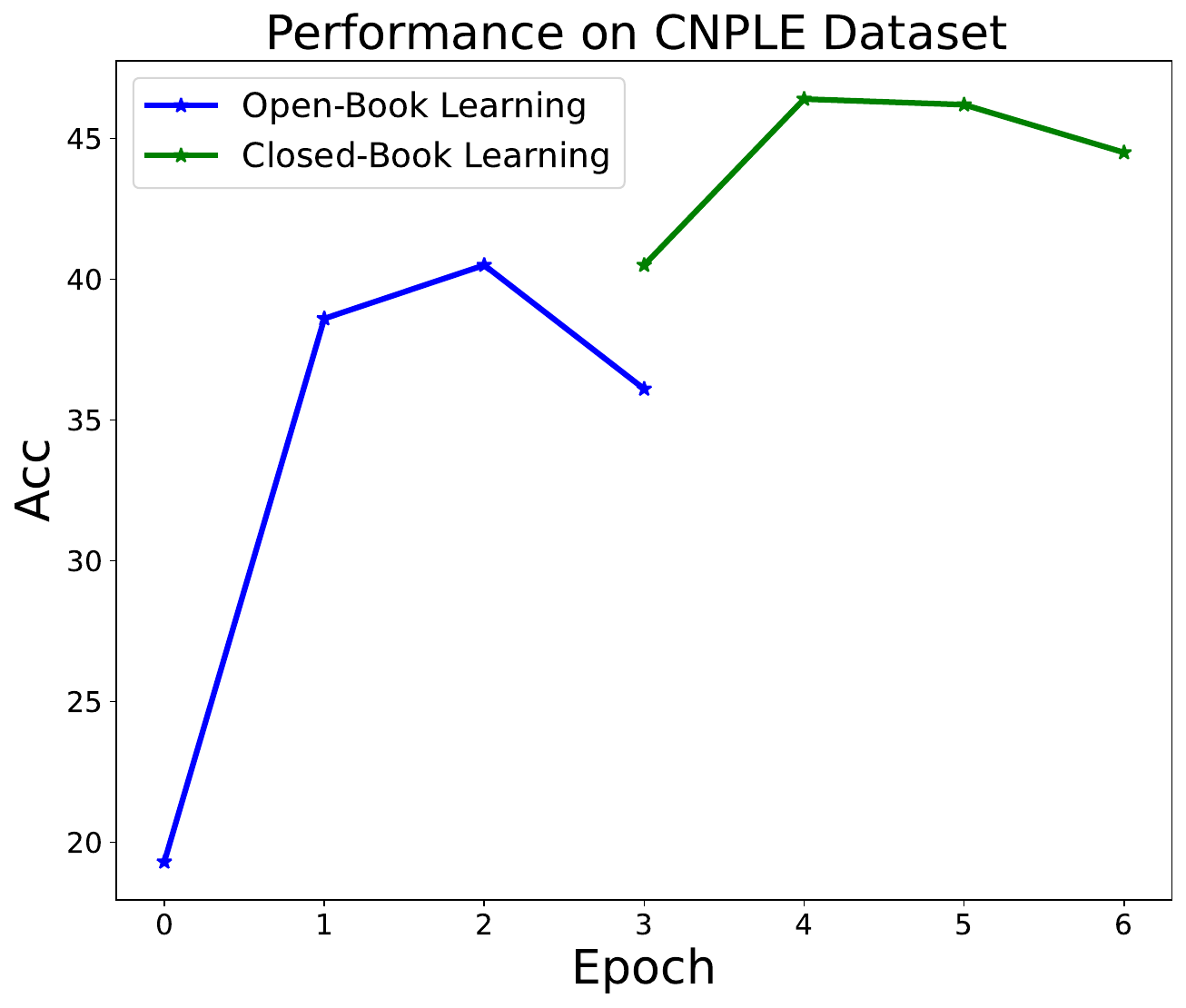}
        \centerline{(d)}
    \end{minipage}
    \begin{minipage}[t]{0.33\linewidth}
        \centering
        \includegraphics[width=\textwidth]{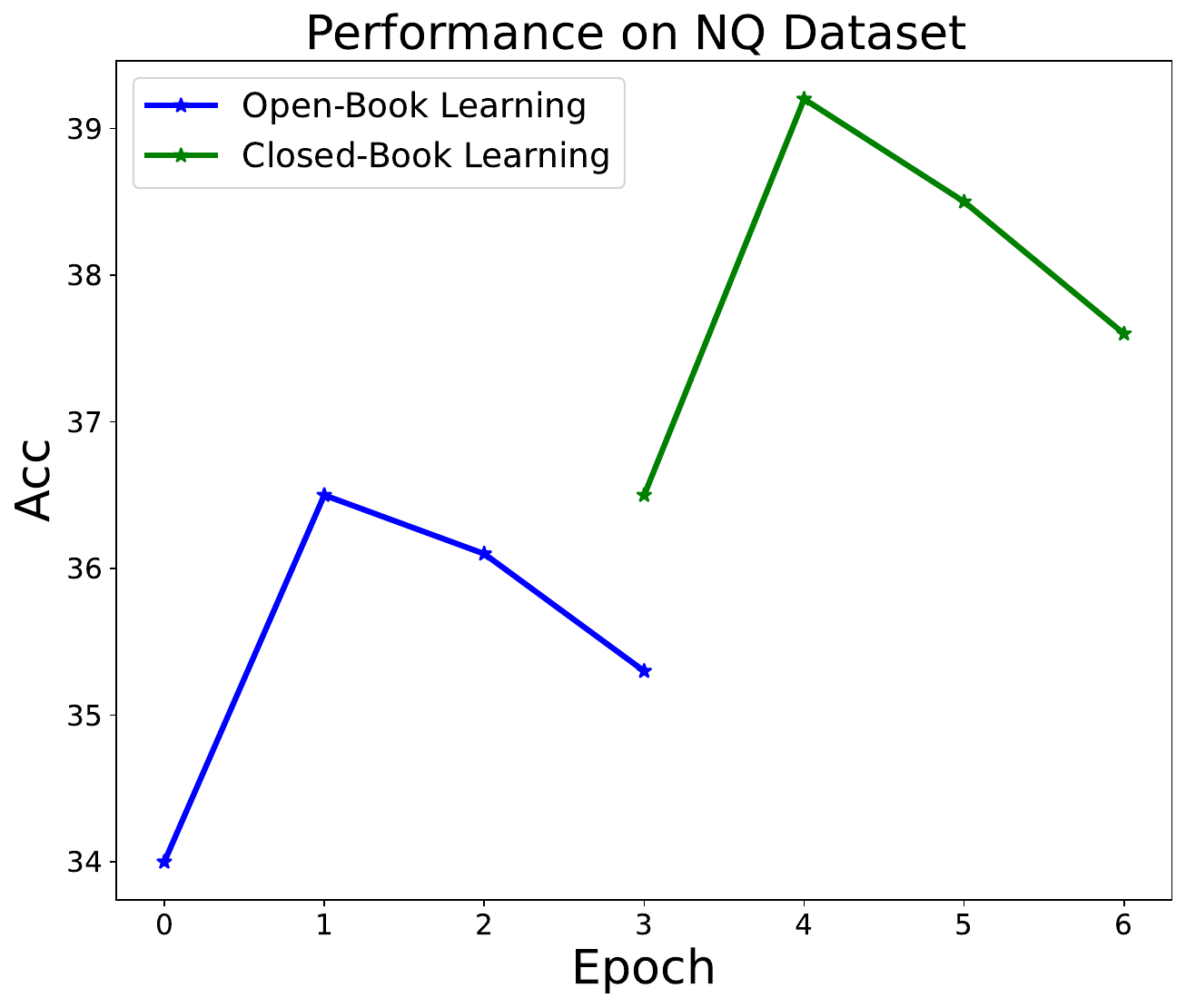}
        \centerline{(e)}
    \end{minipage}
    \caption{
    The performance gap between open-book learning and closed-book learning.
    Epoch 0 stands for the performance of initial model.
    Epochs 4 to 6 represent the 1st, 2nd, and 3rd epochs of closed-book learning, respectively.
    }
\label{fig:not_enough}
\end{figure*}

In this section, we emphasize the point of this paper, that Naive Naive Empirical Risk Minimization (EMR) is not enough, through trend charts on various datasets. 
In Figures~\ref{fig:not_enough}, it can be observed that all Naive EMR methods exhibit clear plateaus, and additional epoch training does not yield higher performance but rather leads to overfitting. The closed-book learning method introduced in the second stage of this paper further enhances the model's knowledge generalization, resulting in improved accuracy for the corresponding tasks, indicating the effectiveness of the knowledge-contrasting approach proposed in this paper.

\section{Mathematical Derivations of AL}
\label{appendix:math_of_AL}
In this appendix, 
we will clarify that our approach is a process of autonomously enhancing knowledge generalization based on knowledge comparison, rather than simply praising or criticizing. We propose the advantages of RL methods in two ways.

\begin{figure*}[ht]
    \begin{minipage}[t]{0.24\linewidth}
        \centering
        \includegraphics[width=\textwidth]{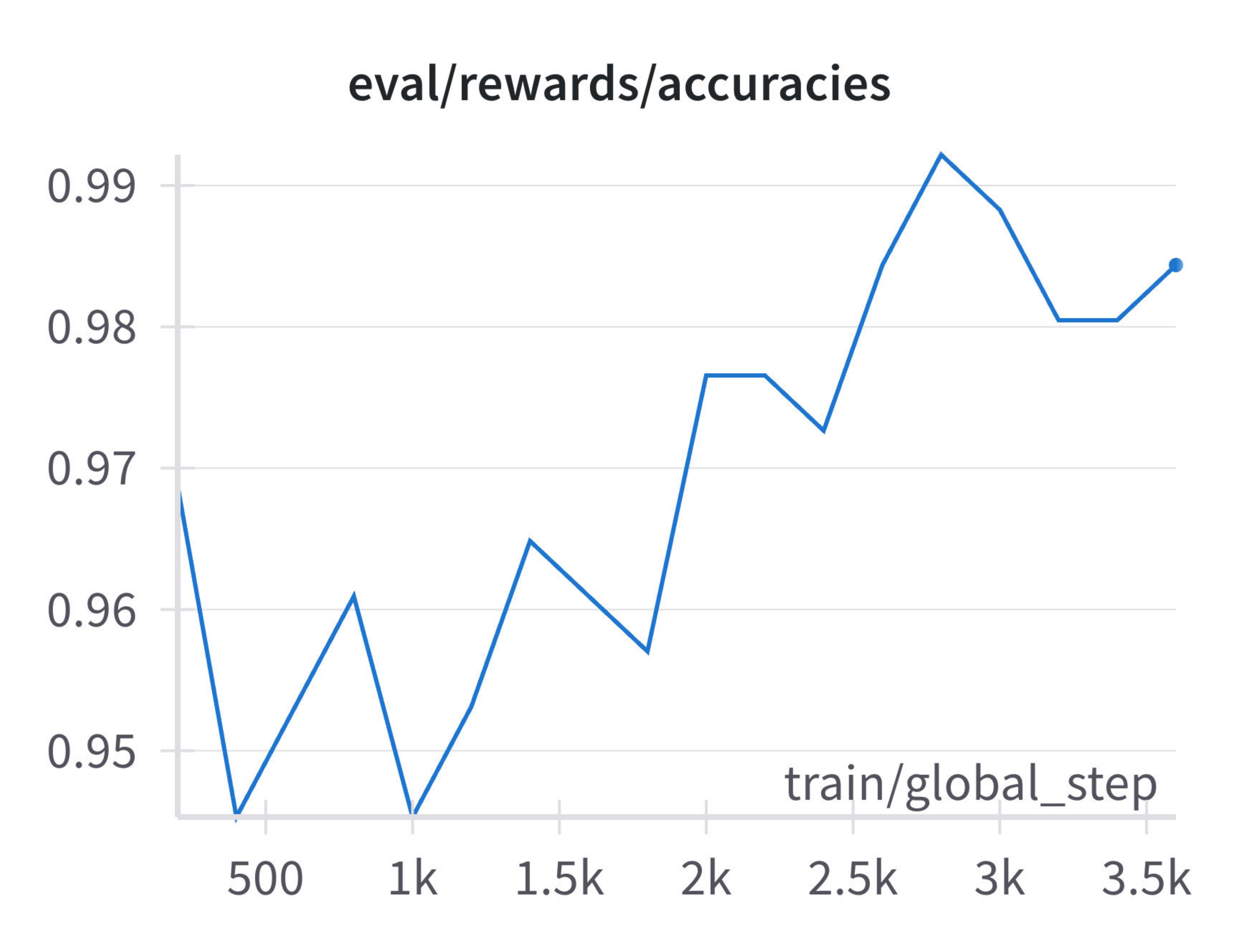}
        \centerline{(a) Reward accuracy}
    \end{minipage}%
    \begin{minipage}[t]{0.24\linewidth}
        \centering
        \includegraphics[width=\textwidth]{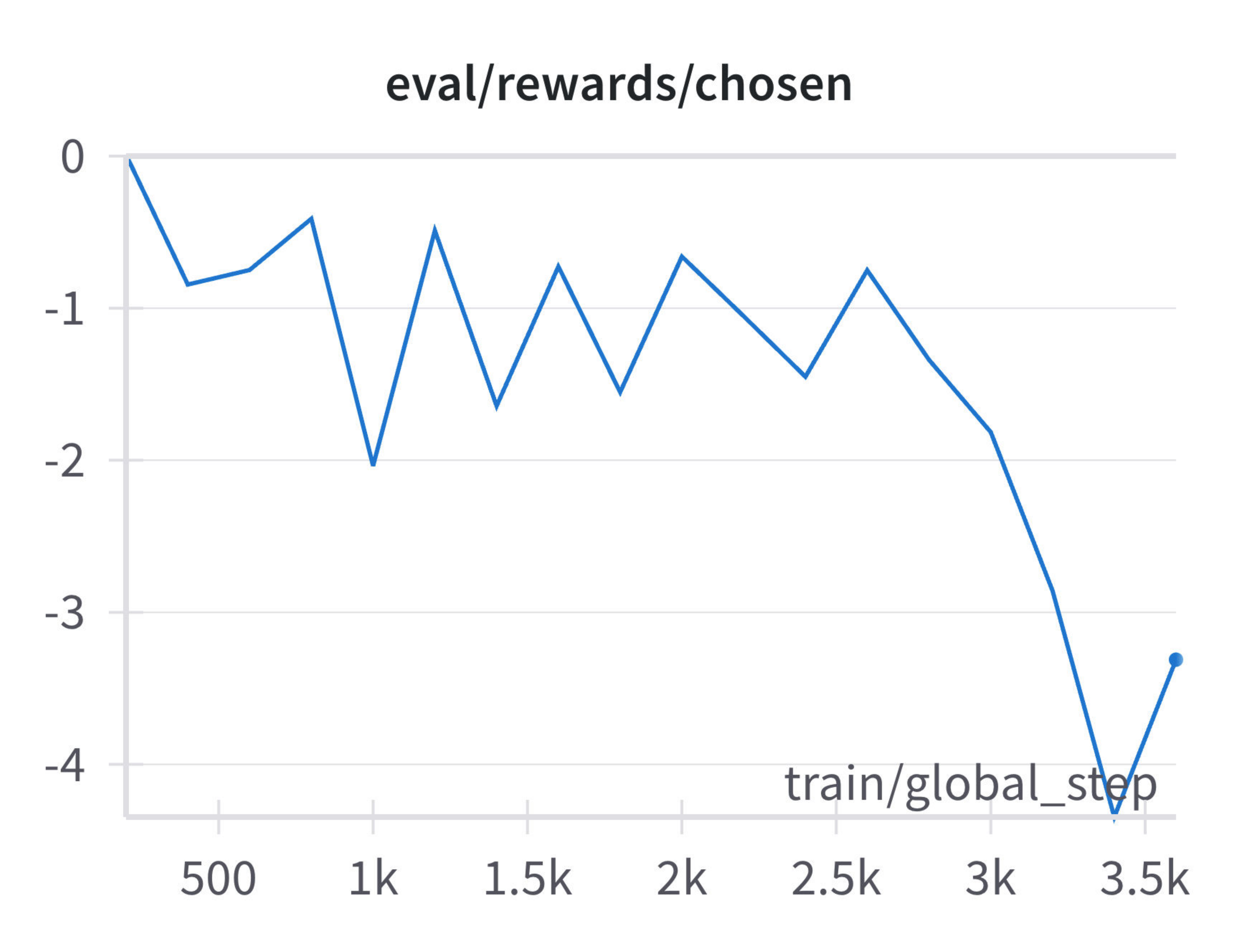}
        \centerline{(b) Reward chosen}
    \end{minipage}
    \begin{minipage}[t]{0.24\linewidth}
        \centering
        \includegraphics[width=\textwidth]{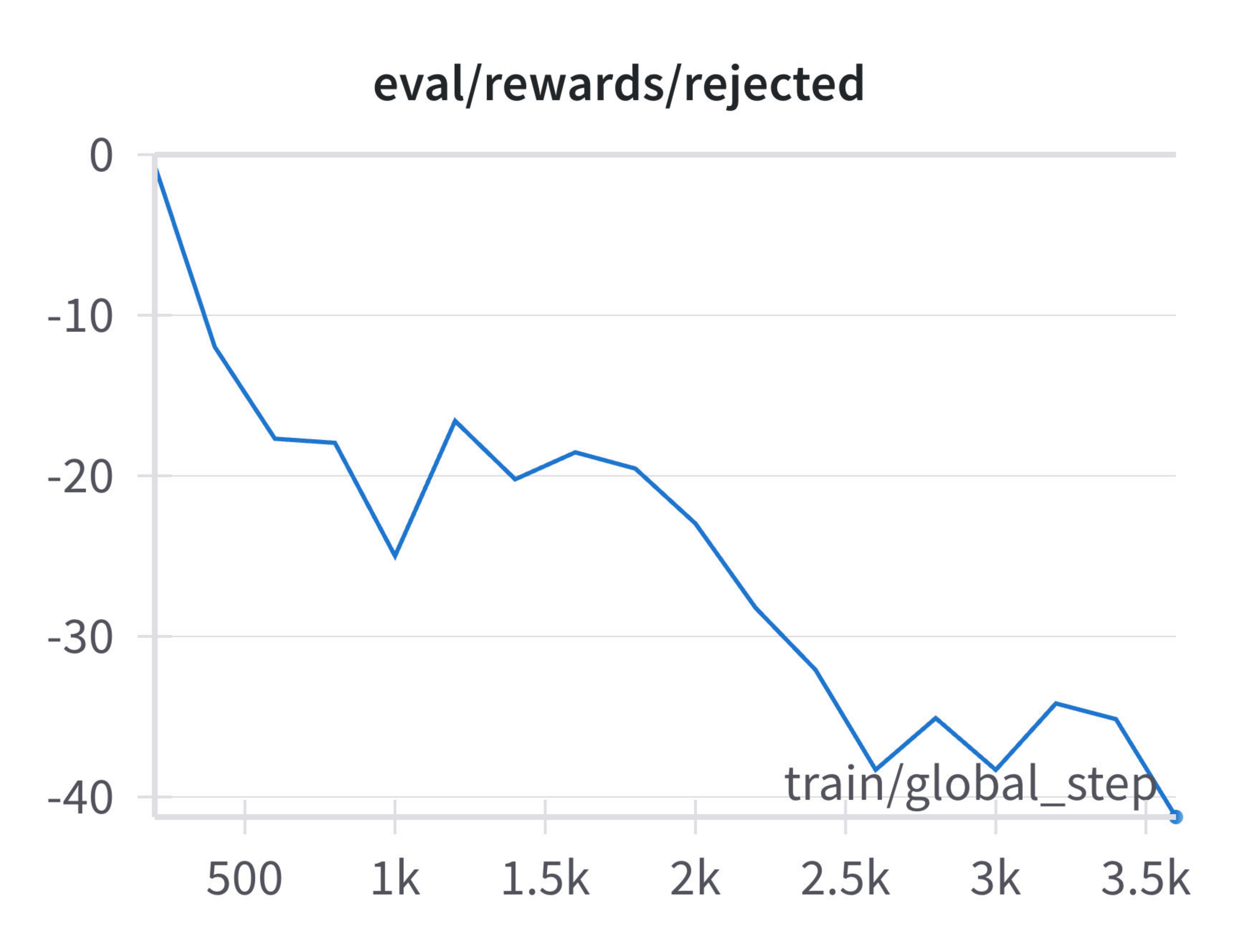}
        \centerline{(c) Reward rejected}
    \end{minipage}
    \begin{minipage}[t]{0.24\linewidth}
        \centering
        \includegraphics[width=\textwidth]{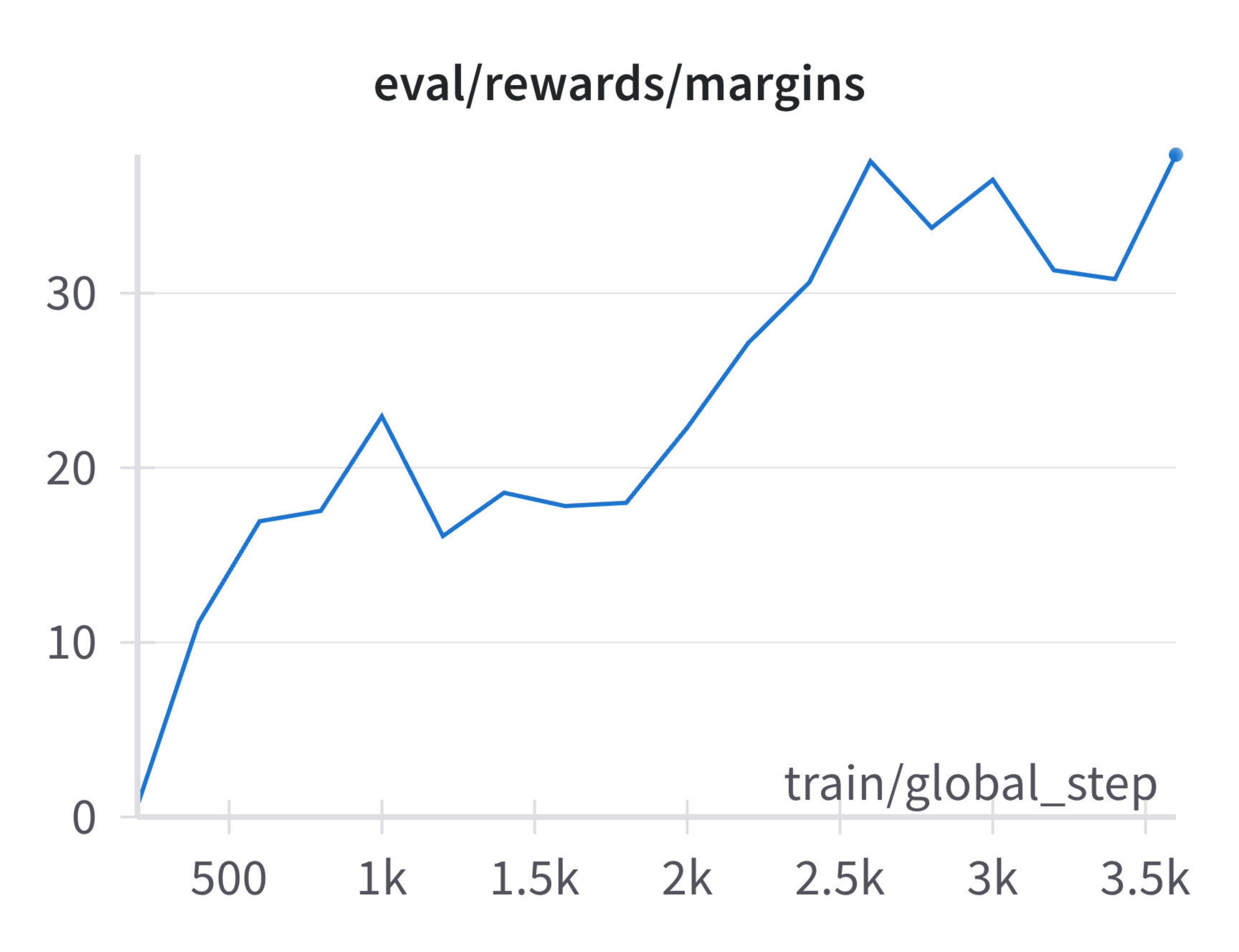}
        \centerline{(d) Reward margins}
    \end{minipage}
    \caption{
    Reward performance on MedQA-cn dataset.
    }
\label{fig:P_metric}
\end{figure*}

First, by \cite{rafailov2023direct} Section 4, the gradient of DPO loss is:
$$\nabla_\theta\mathcal{L}_{DPO} = -\beta E_{(x,y_w,y_l)\sim D}[\sigma(\hat{r}_\theta(x, y_l) - \hat{r}_\theta(x, y_w))(\nabla_\theta\log\pi_\theta(y_w|x) - \nabla_\theta\log\pi_\theta(y_l|x))]$$
where $(x,y_w)$ and $(x,y_l)$ are the chosen and rejected responses, respectively. The updated parameters of the model will move in the direction making the difference $\nabla_\theta\log\pi_\theta(y_w|x) - \nabla_\theta\log\pi_\theta(y_l|x)$ become larger with a weight function $\sigma(\hat{r}_\theta(x, y_l) - \hat{r}_\theta(x, y_w))$, not just increase the log probability of the chosen one and decrease the log probability of the rejected one. Actually in \cite{rafailov2023direct} Appendix C, it has been shown that if we just increase the chosen probability and decrease the rejected probability, the language model will degenerate. Our experiment (Figure \ref{fig:P_metric}) shows that the rewards of chosen and rejected responses can be increase or decrease simultaneously. 

Second, by Equation (4) in \cite{rafailov2023direct}, the optimal solution of the KL-constrained reward maximization objective is:
$$\pi(y|x) = \frac{1}{Z(x)}\pi_{ref}(y|x)\exp(\frac{1}{\beta}r(x,y))$$
for the given reference model $\pi_{ref}$ and reward $r$, where $Z(x)$ is the normalization factor independent of the responses. Hence we can see that the optimal solution is not just choose the best response and ignore all other ones, it is distributed to all responses with the probability determined by the reward function and $\beta$, higher reward leads to higher probability. It can be seen that for two different responses $y_1,y_2$, although there is a better one, but if they are both good enough, that means $r(x,y_1)$ and $r(x,y_2)$ are closed with each other, there probabilities in the optimal distribution will be closed. So the RL methods for the LLM training is not just praising or criticizing, but only depends on their actually rewards. Responses with high reward values will have high probabilities in the end.

\begin{figure}[t]
    
\begin{minipage}[t]{0.49\linewidth}
    \centering
    \includegraphics[width=\textwidth]{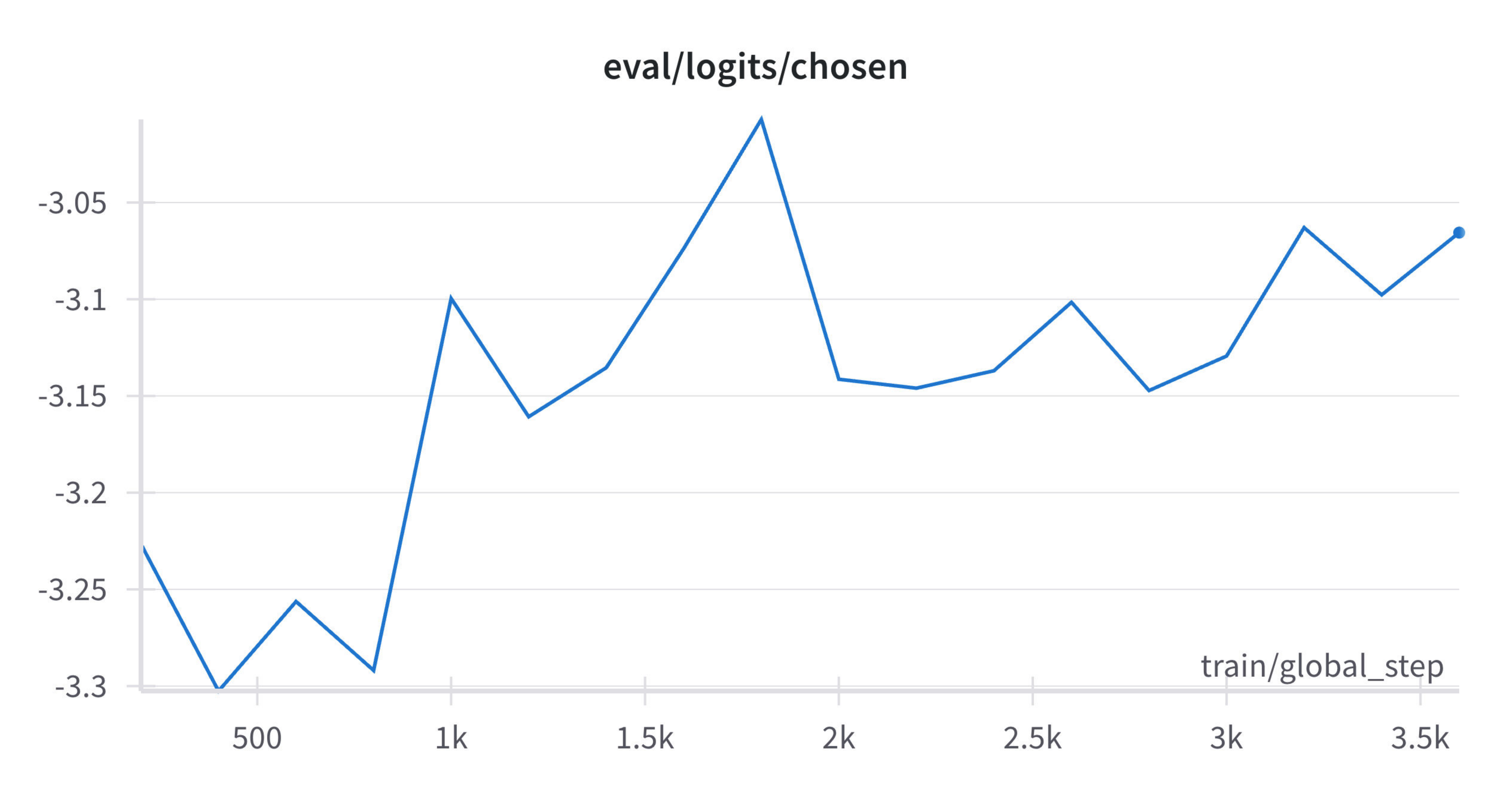}
    \centerline{(a) Logits for the open-book answer.}
\end{minipage}%
\begin{minipage}[t]{0.49\linewidth}
    \centering
    \includegraphics[width=\textwidth]{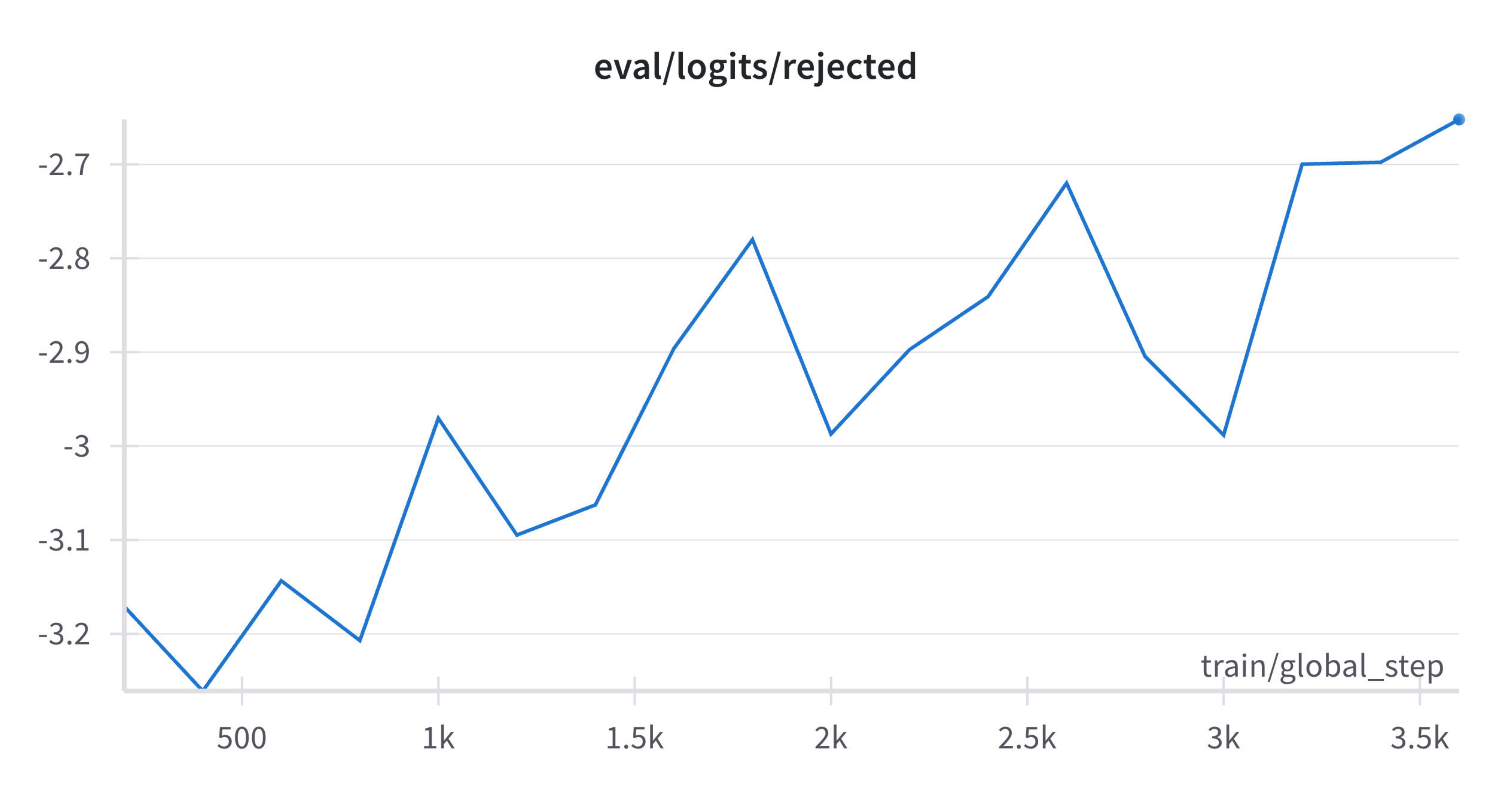}
    \centerline{(b) Logits for the closed-book answer.}
\end{minipage}
\caption{
The trend of logits variation for open-book answers and closed-book answers on the MedQA-cn dataset.
}
\label{fig:logits}
\end{figure}

\section{Demonstrating How Autonomous Learning Works Through Examples}
\label{appendix:case_study}

\begin{figure}[b]
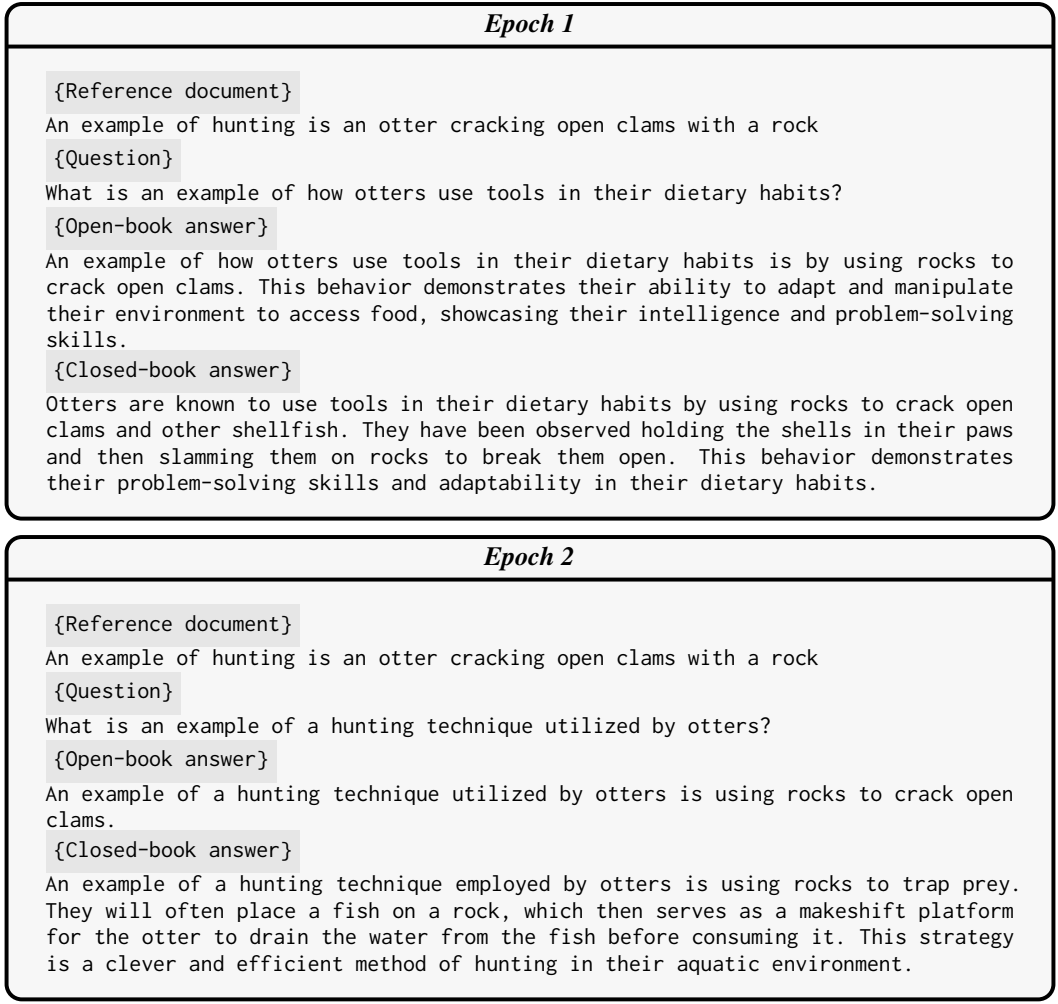

\centering
\label{fig:case_study}

\begin{patient}{\centering \textit{Epoch 1}} 
\colorbox{outerboxcolor}{\strut \{Reference\ document\}}

An example of hunting is an otter cracking open clams with a rock

\colorbox{outerboxcolor}{\strut \{Question\}}

What is an example of how otters use tools in their dietary habits?

\colorbox{outerboxcolor}{\strut \{Open-book answer\}}

An example of how otters use tools in their dietary habits is by using rocks to crack open clams. This behavior demonstrates their ability to adapt and manipulate their environment to access food, showcasing their intelligence and problem-solving skills.

\colorbox{outerboxcolor}{\strut \{Closed-book answer\}}

Otters are known to use tools in their dietary habits by using rocks to crack open clams and other shellfish. They have been observed holding the shells in their paws and then slamming them on rocks to break them open. This behavior demonstrates their problem-solving skills and adaptability in their dietary habits.
\end{patient}

\begin{patient}{\centering \textit{Epoch 2}}
\colorbox{outerboxcolor}{\strut \{Reference\ document\}}

An example of hunting is an otter cracking open clams with a rock

\colorbox{outerboxcolor}{\strut \{Question\}}

What is an example of a hunting technique utilized by otters?

\colorbox{outerboxcolor}{\strut \{Open-book answer\}}

An example of a hunting technique utilized by otters is using rocks to crack open clams.

\colorbox{outerboxcolor}{\strut \{Closed-book answer\}}

An example of a hunting technique employed by otters is using rocks to trap prey. They will often place a fish on a rock, which then serves as a makeshift platform for the otter to drain the water from the fish before consuming it. This strategy is a clever and efficient method of hunting in their aquatic environment.
\end{patient}
\caption{An example to show how Autonomous Learning works.}
\end{figure}

In this appendix, we demonstrate how Autonomous Learning works through some examples. As shown in Figure~\ref{fig:case_study}, we observe that after one epoch of closed-book learning, the closed-Book answer in Epoch 2 aligns better with the learned documents and questions that the closed-book answer in Epoch 1.

\section{Limitations}

Despite its promising performance in knowledge learning, Autonomous Learning has some limitations that must be considered:

\begin{itemize}
    \item \textbf{Limited to models with instruction-following capabilities.:} 
    The method of this paper starts directly from an initial model, which needs to have sufficient instruction-following capabilities to complete both open-book and closed-book answers.
    However, for models that do not possess this instruction-following capability like GPT-2~\cite{radford2019language}, we can use chat models like Llama-2-7b-chat-hf~\cite{touvron2023llama2}, Baichuan 2-Chat-7b~\cite{yang2023baichuan}, ChatGPT~\cite{chatgpt} to simply construct instruction fine-tuning datasets to enable them to master the instruction-following required for Autonomous Learning. 
\end{itemize}

\newtcolorbox{patient}[1]{
        boxrule = 1.5pt,
        fontupper = \small\tt,
        fonttitle = \bf\color{black},
        arc = 5pt,
        rounded corners,
        colframe = black,
        colbacktitle = white!97!black,
        colback = white!97!black,
        title = #1,
}

\begin{figure}[h]
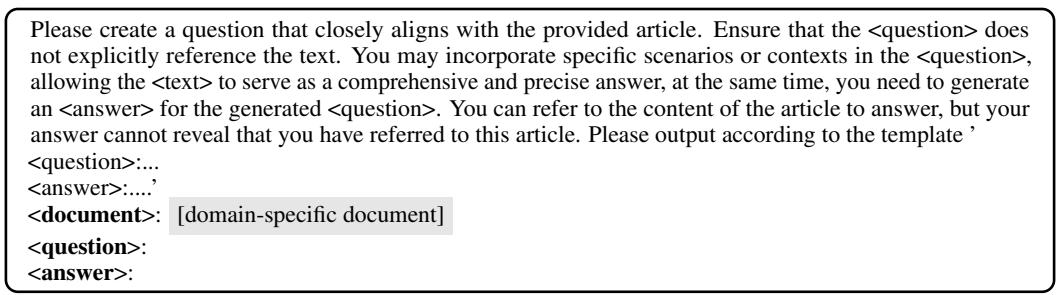

\small
\begin{tcolorbox}[colback=innerboxcolor, colframe=innerboxcolor, colframe=black, boxrule=1pt, arc=4pt, left=6pt, right=6pt, top=1pt, bottom=1pt]
Please create a question that closely aligns with the provided article. Ensure that the <question> does not explicitly reference the text. You may incorporate specific scenarios or contexts in the <question>, allowing the <text> to serve as a comprehensive and precise answer, at the same time, you need to generate an <answer> for the generated <question>. You can refer to the content of the article to answer, but your answer cannot reveal that you have referred to this article. Please output according to the template '

<question>:...

<answer>:....'

<\textbf{document}>: \colorbox{outerboxcolor}{[domain-specific document]}

<\textbf{question}>: 

<\textbf{answer}>: 
\end{tcolorbox}
\caption{
The prompt for question generation.
\colorbox{outerboxcolor}{\rule{0pt}{8pt}[domain-specific document]} refers to a document in the domain-specific pre-training corpora.}
\label{fig:prompt_dqa}
\end{figure}

\begin{figure}[h]
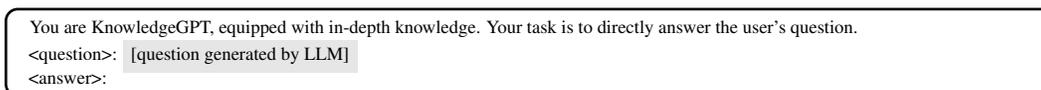

\scriptsize
\begin{tcolorbox}[colback=innerboxcolor, colframe=innerboxcolor, colframe=black, boxrule=1pt, arc=4pt, left=6pt, right=6pt, top=1pt, bottom=1pt]
You are KnowledgeGPT, equipped with in-depth knowledge. Your task is to directly answer the user's question. 

<question>: \colorbox{outerboxcolor}{[question generated by LLM]}
    
<answer>: 
\end{tcolorbox}
\caption{
The prompt for the answer generation of $\purple{\mathcal{QA}}$.
\colorbox{outerboxcolor}{\strut[question generated by LLM]]} is the previously text-derived query in Figure~\ref{fig:prompt_dqa}. 
}
\label{fig:prompt_qa}
\end{figure}

\begin{table}[h]
\caption{
An example of our $\purple{\mathcal{QA}}$ instruction tuning data.
}
\footnotesize
\begin{tabular}{p{0.95\linewidth}}
\toprule
{\normalsize \textbf{<|system|>}}\newline
You are KnowledgeGPT, equipped with in-depth knowledge. 
Your task is to directly answer the user's question. \newline
{\normalsize \textbf{<|user|>}}\newline
[question] What is an example of a film that combines comedy with a critique of human vices through interwoven stories, highlighting the impact of greed among others?\newline
{\normalsize \textbf{<|assistant|>}}\newline
[answer] An example of a film that skillfully combines comedy with a critique of human vices, specifically focusing on the theme of greed among others...\\
\bottomrule
\end{tabular}
\label{tab:instruction_template_qa}
\end{table}
\end{document}